\documentclass{article}

\usepackage[preprint]{neurips_2024}
\usepackage{multirow}

\usepackage[utf8]{inputenc} 
\usepackage[T1]{fontenc}    
\usepackage{hyperref}       
\usepackage{url}            
\usepackage{booktabs}       
\usepackage{amsfonts}       
\usepackage{nicefrac}       
\usepackage{microtype}      
\usepackage{xcolor}         
\usepackage{amsmath}
\usepackage{amssymb}
\usepackage{mathtools}
\usepackage{amsthm}
\usepackage{multirow}
\usepackage{caption}
\usepackage{pifont}
\usepackage{multirow}
\usepackage{hyperref}
\usepackage{url}
\usepackage{graphicx}
\usepackage{amsmath}
\usepackage{amssymb}
\usepackage{booktabs}
\usepackage{amsfonts}
\usepackage{amsfonts}
\usepackage{times}
\usepackage{epsfig}
\usepackage{graphicx}
\usepackage{amsmath}
\usepackage{amssymb}
\usepackage{graphicx}
\usepackage{amsmath}
\usepackage{amssymb}
\usepackage{bbm}
\usepackage{booktabs}
\usepackage{multirow}
\usepackage{caption}
\usepackage{pifont}
\usepackage{multirow}
\usepackage{subcaption}
\usepackage{amsmath}
\DeclareMathOperator*{\argmax}{arg\,max}

\DeclareMathOperator*{\weight}{weight}
\newcommand{\cmark}{\ding{51}}%
\newcommand{\xmark}{\ding{55}}%
\renewcommand{\cite}[1]{\citep{#1}}

\title{BaFTA: Backprop-Free Test-Time Adaptation For Zero-Shot Vision-Language Models}

\author{
    \textbf{Xuefeng Hu}$^{1}$, \space
    \textbf{Ke Zha}$^{2}$, \space
    \textbf{Min Sun}$^{2}$, \space
    \textbf{Albert Chen}$^{2}$, \space
    \textbf{Cheng-Hao Kuo}$^{2}$ and
    \textbf{Ram Nevatia}$^{1}$ \AND
    $^{1}$ University of Southern California \And
    $^{2}$ Amazon \AND
    $^{1}$ \texttt{\{xuefengh,nevatia\}@usc.edu} \And
    $^{2}$ \texttt{\{kezha,minnsun,aycchen,chkuo\}@amazon.com}
}

\begin{document}

\maketitle

\begin{abstract}
    Large-scale pretrained vision-language models like CLIP have demonstrated remarkable zero-shot image classification capabilities across diverse domains. To enhance CLIP's performance while preserving the zero-shot paradigm, various test-time prompt tuning methods have been introduced to refine class embeddings through unsupervised learning objectives during inference. However, these methods often encounter challenges in selecting appropriate learning rates to prevent collapsed training in the absence of validation data during test-time adaptation. 
    In this study, we propose a novel backpropagation-free algorithm BaFTA for test-time adaptation of vision-language models. Instead of fine-tuning text prompts to refine class embeddings, our approach directly estimates class centroids using online clustering within a projected embedding space that aligns text and visual embeddings. We dynamically aggregate predictions from both estimated and original class embeddings, as well as from distinct augmented views, by assessing the reliability of each prediction using Rényi entropy.
    Through extensive experiments, we demonstrate that BaFTA consistently outperforms state-of-the-art test-time adaptation methods in both effectiveness and efficiency.
\end{abstract}

\section{Introduction}
    The emergence of large-scale pre-trained ``foundation'' vision-language models (VLM) such as CLIP~\cite{radford2021learning} and ALIGN~\cite{jia2021scaling} has marked a paradigm shift in the field of computer vision. These models have demonstrated promising capacity for open-world generalization, and can be easily applied on novel tasks beyond the original training data. By harnessing an unified visual-text embedding space, VLMs can perform zero-shot image classification on novel concepts by translating the category names into this shared representation space as classification weights. 

In the pursuit to further enhancing performance of such VLMs, various adaptation and fine-tuning techniques have emerged to bridge the domain gap over downstream tasks. For instance, \citet{zhou2022coop}, \citet{zhou2022cocoop} and \citet{samadh2023align} fine-tune text prompts for VLMs, tailoring them to specific downstream tasks with few-shot adaptation. Moreover, in the realm of zero-shot classification, numerous approaches have been proposed to boost VLM performance without requiring labeled data. For example, \citet{xuefeng2023reclip} and \citet{tanwisuth2023pouf} improve CLIP through source-free unsupervised adaptation using unlabeled test examples, while \citet{udandarao2022sus}, \citet{guo2023calip} and \citet{ge2023improving} enhance the CLIP with training-free methods by utilizing the class hierarchy, gradient-based visual attention or external modules such as stable diffusion.  Furthermore, test-time prompt tuning algorithms, such as \citet{shu2022tpt}, \citet{park2023robust} and \citet{ma2024swapprompt}, propose to refine learnable text prompts at inference time through the optimization of an unsupervised objective using augmentations, which lead to improved model accuracy. Recently, we also noted a contemporaneous work by~\citet{karmanov2024efficient} on training-free test-time adaptation for VLMs. The online and label-free nature of these test-time prompt tuning methods underscores their practicality and versatility as effective means to enhance the performance of CLIP, especially within the context of zero-shot classification. 

However, as pointed out by \citet{niu2023towards}, test-time adaptation methods such as \citet{wang2020tent}, \citet{liang2023comprehensive} as well as methods like TPT, often encounter the intricate challenge of determining an optimal learning rate in the absence of validation data to avoid collapsed training. Striking the right balance is crucial—achieving maximum improvement while simultaneously safeguarding against the model's instability during test-time adaptation. 

To address the challenge of stable and efficient test-time adaptation, we propose the Backpropagation-Free Test-time Adaptation (BaFTA) algorithm. BaFTA eliminates the need for backpropagation and reliance on labeled examples, offering a stable and efficient method that achieves strong overall accuracy. BaFTA directly refines class embeddings within the unified visual-text embedding space of CLIP by leveraging neighboring information among test example visual embeddings with an online clustering algorithm. This novel approach mitigates the risk of collapsed training, leverages the natural clustering of high-quality visual embeddings from CLIP, and avoids the instability from noisy test-time self-supervised training.

To further enhance the performance of online clustering predictions, we propose two key designs. First, recognizing that clustering-based predictions can be influenced by the biased distribution of test examples, we combine these predictions with standard predictions derived from randomly augmented views of the test examples. We use Rényi Entropy to evaluate the reliability of these predictions, ultimately arriving at an aggregated prediction that benefits from the strengths of both approaches while ensuring accuracy and robustness. Second, building upon insights from \citet{xuefeng2023reclip}, we execute the online clustering algorithm in a projected embedding space, which helps alleviate the disparity between CLIP’s visual and text embeddings, leading to improved clustering outcomes.

Our key contributions are summarized as follows:
\begin{itemize}
    \item \textbf{Stable and Efficient Test-Time Adaption via Backpropagation-Free Strategy}. We propose and validate the feasibility of a stable and efficient backpropagation-free algorithm achieving strong overall performance in VLM test-time adaptation. By leveraging the natural clustering of projected visual embeddings, BaFTA directly estimates classification centers and avoids the instability from test-time backpropagation training.
    \item \textbf{Stable online clustering via Renyi entropy aggregation}. Naive online clustering often suffers from biased online assignments. To address this challenge, we propose a novel Rényi Entropy Aggregation mechanism to dynamically combine predictions from both text and clustering predictions based on their reliability.
    \item \textbf{Extensive experiments and strong results}. Through extensive experiments, we validate the effectiveness of BaFTA and its novel components, significantly improving the zero-shot classification accuracy of pre-trained vision-language models at inference time with considerably faster speed.
\end{itemize}

\section{Background}
    \label{sec:bg}

\begin{table}[!ht]
\centering
\resizebox{\textwidth}{!}{
\setlength{\tabcolsep}{0.5em}
\begin{tabular}{ccccccccccccc} 
         & \rotatebox{90}{Cars} & \rotatebox{90}{Caltech101} & \rotatebox{90}{DTD}  & \rotatebox{90}{EuroSAT} & \rotatebox{90}{FGVC} & \rotatebox{90}{Food101} & \rotatebox{90}{Flower102} & \rotatebox{90}{Pets} & \rotatebox{90}{UCF101} & \rotatebox{90}{SUN397} & \rotatebox{90}{ImageNet} \\ \midrule
CLIP (RN50)~\textit{Zero-Shot}       & 55.8          & 82.1          & 41.7          & 41.1          & 19.3          & 81.1          & 65.9          & 85.4          & 63.6          & 59.6          & 59.6          \\
CLIP (RN50)~\textit{Linear-Eval}     & \textbf{78.3} & \textbf{89.6} & \textbf{76.4} & \textbf{95.2} & \textbf{49.1} & \textbf{86.4} & \textbf{96.1} & \textbf{88.2} & \textbf{81.6} & \textbf{73.3} & \textbf{73.3} \\ \midrule
CLIP (ViT-B/16)~\textit{Zero-Shot}   & 65.6          & 89.3          & 46.0          & 54.1          & 27.1          & 89.2          & 70.4          & 88.9          & 69.8          & 65.2          & 68.6          \\
CLIP (ViT-B/16)~\textit{Linear-Eval} & \textbf{86.7} & \textbf{94.7} & \textbf{79.2} & \textbf{97.1} & \textbf{59.5} & \textbf{92.8} & \textbf{98.1} & \textbf{93.1} & \textbf{88.4} & \textbf{78.4} & \textbf{80.2}
 \\ \bottomrule \\
\end{tabular}}
\caption{Zero-Shot v.s. Linear Evaluation top-1 accuracy reported by CLIP~\cite{radford2021learning}. Linear Evaluation protocol assesses the quality of visual embeddings by training a fully-supervised linear classifier over the frozen visual embeddings. This Linear Evaluation result implies: \textbf{1)} the zero-shot performance of CLIP are largely limited by the quality of zero-shot classifier, i.e, the text embeddings of class names; \textbf{2)} The native visual embeddings of CLIP get classified well with a linear classifier, which suggests the distinctiveness of visual embeddings across target classes, and leads to an opportunity to leverage the neighboring relationships to enhance test-time performance.} 
\label{lineareval}
\end{table}



In this section, we revisit the large-scale pre-trained vision language model CLIP~\cite{radford2021learning} and test-time prompt tuning algorithm TPT~\cite{shu2022tpt} for the necessary background and formulation before we introduce our method in Section~\ref{sec:poc}. 

\textbf{Zero-Shot Image Classification with VLM.} A pre-trained vision-language model such as CLIP consists of two parallel components $M=\{M_v, M_t\}$ where $M_v$ is the visual encoder and $M_t$ is the text encoder. Given test images $D^{test}=\{x_i\}_{i=1}^I$ and target class names $C=\{c_j\}_{j=1}^J$ , the pre-trained vision-language model $M$ performs zero-shot classification by generating the adaptive classification weights from text embeddings of the target class names $t_j = M_t(\theta_0(c_j))$ for $j\in\{1,2,...,J\}$, where $\theta_0$ is the text prompt template such as ``\texttt{a photo of a \{class name\}}'' that warpped the class names $c_j$ into full sentences $\theta_0(c_j)$. 
To further improve the quality of the text embeddings, CLIP provides lists of templates $\{\theta_z\}_{z=1}^Z$ to align the text embeddings with the distribution of real captions used in pre-training,  and generates the text embeddings for each class by taking the average of these templates,
$$
    t_j = \frac{1}{Z}\sum_{z=1}^Z M_t(\theta_z(c_j)).
$$
 Then, the prediction $y_i$ can be obtained by selecting the class $j$ whose text embedding $t_j$ has the highest cosine similarity with its visual embedding $M_v(x_i)$, i.e.,
$$
    y_i = \argmax_{j} \langle \frac{M_v(x_i)}{\lVert M_v(x_i) \rVert}, \frac{t_j}{\lVert t_j \rVert}\rangle.
$$

\textbf{Test-Time Prompt Tuning for VLM.} 
To further enhance the zero-shot generalization ability of vision language model $M$, TPT proposes to learn an adaptive text template $\theta$ at inference time. For each test example $x_i$, TPT first prepares a mini-batch of random augmented views $\{x_i^1, x_i^2, ..., x_i^B\}$ and performs a single step gradient descent to optimize the entropy minimization loss over the high-confidence predictions among the augmented views,
$$
    \theta_i = \theta_0 - \delta \nabla_{\theta} \left(\sum_{b=1}^B \mathbbm{1}[H(M(x_i^b) < \tau] H(M(x_i^b)) \right)\rvert_{\theta=\theta_0}
$$
where $H(\cdot)$ is the entropy function, $\tau$ is the entropy threshold for high-confidence augmented view selection, and $\delta$ is the learning rate. $M(x_i^b) = softmax\left(\left[M(x_i^b, c_j, \theta)\right]_{j=1}^J\right)$ is the estimated probability distribution of augmented view $x_i^b$ over target classes $c_1,...,c_j$, with $M(x_i^b, c_j, \theta)=\left\langle \frac{M_v(x_i^b)}{\lVert M_v(x_i^b) \rVert}, \frac{M_t(\theta(c_j)}{\lVert M_t(\theta(c_j) \rVert}\right\rangle$ as the cosine-similarity between visual embedding $M_v(x_i^b)$ and text embedding $M_t(\theta(c_j))$.   
Then, with adapted text prompt $\theta_i$, TPT produces the prediction for test example $x_i$ with the original image:
\begin{equation}
    y_i = \argmax_{j} M(x_i^b, c_j, \theta) \label{entropy_conf}
\end{equation}

\section{Method}
    \label{sec:mc}
\begin{figure*}[ht]
    \centering
    \includegraphics[width=\textwidth]{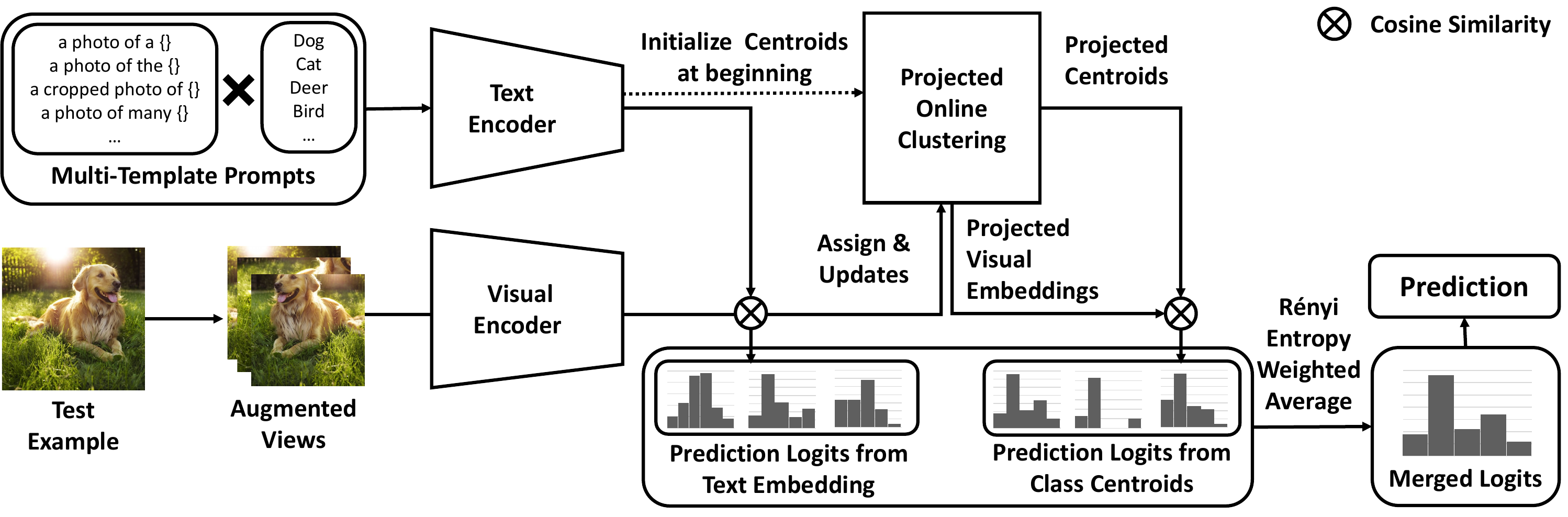}
    \caption{Overview of the Backpropagation-Free Test-Time Adaptation algorithm BaFTA. Instead of prompt-tuning, we employ online clustering to directly estimate class embeddings in a projection space that aligns visual and text embeddings. The class centroids are initialized with text embeddings of class names, and updated incrementally with online test examples assigned to the class. For each test example, we generate two sets of predictions. The first set measures cosine similarity between visual embeddings of augmented views and class name text embeddings. The second set measures cosine similarity between visual embeddings and online-clustering centroids. Predictions are aggregated with reliability estimated by Rényi Entropy for final results. }
    \label{fig:overall}
\end{figure*}

\label{sec:poc}
As investigated by \citet{niu2023towards}, test-time adaptation algorithms frequently encounter the challenge regarding the appropriate choice of the learning rate in absence of validation data during unsupervised training. On one hand, opting for a small learning rate will restrict the enhancement of the model. On the other hand, employing a large learning rate can be risky in triggering the potential model collapse. TPT adopts a relatively large learning rate to expedite improvement, but chooses to restart from the original model for each test example to prevent the potential model collapse. 

In this work, we present an novel backpropagation-free solution which directly refines the class embeddings in the aligned visual-text embedding space instead of in the prompt embedding space. BaFTA performs \textbf{Ba}ckpropagation-\textbf{F}ree \textbf{T}est-time \textbf{A}daptation for Vision-Language Models, and brings three major advantages over the test-time prompt tuning methods like TPT:

\begin{itemize}
    \item BaFTA avoids the use of back-propagation to update model weights. As a result, it achieves significantly faster inference speed while greatly reduces risk of causing model collapse during unsupervised adaptation.
    \item In contrast to the test-time adaptation algorithms like TPT that require frequent restart to prevent model collapse, BaFTA possesses the capability to scale as more test examples become available, and to leverage from the relationships between neighboring examples. 
    \item BaFTA can leverage the multi-template prompts provided by CLIP to enhance text embedding quality. In contrast, prompt-tuning methods are constrained to single-template prompts due to computational costs.
\end{itemize}

In the following sections, we first present the motivation and primary concepts behind the estimation of class embeddings using projected online clustering at inference time in Section~\ref{sec:oc}. Subsequently, we dive into the discussion of the key findings that enhance the performance of online clustering, as elaborated in Section~\ref{sec:renyi}. Finally, we present a comprehensive overview of the BaFTA in Section~\ref{sec:alg}.

\subsection{Estimate Class Embedding with Online Clustering}
\label{sec:oc}

As shown in Table~\ref{lineareval}, CLIP generates discriminative visual embeddings on various datasets, but the zero-shot classification performance is often limited by the imprecise text embeddings generated from uninformative class names. For example, FGVC Aircraft~\cite{fgvc} uses codenames such as \texttt{707-320} and \texttt{A300-B4} as class names, which are hardly informative for CLIP to generate proper embeddings to capture the visual difference between classes. 

Conversely, the results of linear evaluation suggest that the visual embeddings from CLIP exhibit a high degree of distinctiveness among target classes, enabling the linear classifier to attain remarkable classification accuracy. This finding opens up an opportunity to leverage the neighboring information within these visual embeddings to further enhance classification performance.

Given a set of visual embeddings $\{v_i|v_i=M_v(x_i)\}_{i=1}^I$ come in order at inference time, we can obtain a set of cluster centroids $w_j$ as class embeddings using the online clustering algorithm ~\citet{barbakh2008online}: 
\begin{align}
    w_j &= \frac{t_j}{\lVert t_j \rVert}  \\ 
    w_{y_i} &= \frac{k_{y_i}w_{y_i} + v_i}{\lVert k_{y_i}w_{y_i} + v_i\rVert} \label{online_cluster_eq} \\ 
    k_{y_i} &= k_{y_i} + 1  
\end{align}
where the cluster centroids $\{w_j\}$ are initialized with the text embeddings $\{t_j\}$, and corresponding centroid $w_{y_i}$ will be averaged with the online examples $v_i$ based on its prediction $y_i$ and the current cluster size $k_{y_i}$.


\subsection{Visual Text Alignment}
\label{sec:proj}

While VLMs aim to establish an unified embedding space for both visual and text modalities, recent research studies conducted by~\citet{liang2022mind},~\citet{tanwisuth2023pouf} and~\citet{xuefeng2023reclip} have suggested that contrastive pre-trained models might still exhibit a notable disparity between their visual and text embeddings. \citet{xuefeng2023reclip} introduces a simple yet effective projection-based alignment method. This method effectively removes the classification-agnostic information that is inherent in both visual and text modalities. As a result, it efficiently aligns the visual and text embeddings, leading to the advantages of enhanced embedding distribution and clustering characteristics.

Assuming a classification task with $J$ classes, where the text embeddings are denoted as $T = [t_1,...,t_J]$ with $t_j=M_t(c_j)$. Using the singular value decomposition $$ U, S, V = svd(T)$$
we obtain $U=[e_1, e_2, ..., e_J]$ as the orthonormal basis of the span of $T$, that defines a matrix $P=UU^\top$. This matrix projects embeddings onto the span of $T$ and removes the class-agnostic information irrelevant to classification. Additionally, the principle axis $e_1$ within the outer space basis $U$ represents where $\{t_1,...,t_J\}$ overlap the most. By removing $e_1$, the text embeddings are separated from each other, which in turn distances the visual embeddings of different classes.
Together with feature re-normalization, \citet{xuefeng2023reclip} defines the projection function $P^*$ with 
\begin{align}
    P^*(x):= \frac{U'U'^\top x}{\lVert U'U'^\top x\rVert} & & U'= [e_2, e_2, ..., e_J]
    \label{proj_eq}
\end{align}

\subsection{Prediction Aggregation with Rényi Entropy}
\label{sec:renyi}
The online clustering algorithm presented in Section~\ref{sec:oc} yields accurate estimations of the embedding centroids for classes that have a sufficient quantity of seen test examples. However, when it comes to classes with only a limited number of examples, the estimations of embedding centroids can become notably biased. In datasets featuring a large number of classes like ImageNet1k~\cite{deng2009imagenet}, certain categories might remain unassigned or have very few examples assigned to them until the adaptation process concludes. This situation reduces the reliability of centroid estimation for these classes. Consequently, it becomes imperative to implement a mechanism for filtering out predictions with low reliability.

On the other hand, we follow TPT~\cite{shu2022tpt} to leverage random augmentations to improve the prediction quality on test examples. For each test example $x_i$, we prepare $B$ augmented views $\{x_i^1,...,x_i^B\}$, which result in a $B$ distinct predictions $\{p_{i}^1,...,p_{i}^B\}$ that also requires to be filtered and to preserve the reliable ones.  As described in Equation~\ref{entropy_conf}, TPT selects the predictions $p_{i}^b$ by thresholding their entropy $H(p_{i}^b) > \tau$, as the high entropy predictions tend to be more confident. 

On the contrary, we draw inspiration from a study from the area of speech recognition~\citet{laptev2023fast} and opt for Rényi Entropy to estimate the reliability of each prediction. This decision is motivated by the observed stronger correlation between Rényi Entropy and prediction accuracy, as indicated in the study. For each test example $x_i$, we generate regular predictions $p_{i}^b$ by calculating the softmax-cosine similarity between visual embedding $v_i^b$ and text embedding $t_j$:  
\begin{equation}
    p_{i}^b = softmax\left(\left[cos(v_i^b, t_j)\right]_{j=1}^J\right), \label{pd1}
\end{equation}
    
and also online-clustering predictions $p_{i}^b$ by comparing $v_i^b$ with the class centroids $w_j$:
\begin{equation}
    \hat{p_{i}^b} = softmax\left(\left[cos(P^*(v_i^b), w_j)\right]_{j=1}^J\right). \label{pd2}
\end{equation}
Note that we use projected visual embeddings $P^*(v_i^b)$ to calculate $\hat{p_{i}^b}$, because $w_j$ are calculated in the projection space. Then, we estimate the reliability of each prediction $p$ with the negative Rényi Entropy:
\begin{equation}
    Re(p) = \frac{1}{\alpha-1} \log \sum_{j=1}^J (p[j])^{\alpha} \label{eq:renyi}
\end{equation}

Finally, we aggregate the predictions $\{p_i^b\}$ and $\{\hat{p_i^b}\}$ with their Rényi entropy as the weight:
\begin{align}
    \Tilde{p_i} &= \frac{\beta}{R_1}\left(\sum_{b=1}^B Re(p_i^b) p_i^b\right) + \frac{1}{R_2}\left(\sum_{b=1}^B Re(\hat{p_i^b}) \hat{p_i^b}\right) \nonumber \\ 
                &= \beta\frac{p_i}{R_1} + \frac{\hat{p_i}}{R_2} \label{eq:renyi2}
\end{align}
where $\beta$ further balances the weight between text embedding predictions and online clustering predictions, and $R_1=(1+\beta)\sum_{b=1}^B Re(p_i^b), R_2=(1+\beta)\sum_{b=1}^B Re(\hat{p_i^b})$ are the normalization factor to ensure $\Tilde{p_i}$ sums to 1.

\subsection{Overview of BaFTA}
\label{sec:alg}
We demonstrate the overview of BaFTA in Figure~\ref{fig:overall}. Instead of employing prompt-tuning, which entails back-propagation and the risk of potential model collapse during unsupervised training, BaFTA takes a backpropagation-free approach. We directly refine the class embeddings with online clustering (as detailed in Section~\ref{sec:oc}) in a projection space that aligns the visual and text embeddings (as detailed in Section~\ref{sec:proj}). 
For each test instance, BaFTA generates two sets of predictions. The first set follows the standard contrastive VLM classification protocol, measuring cosine similarity between visual embeddings of augmented views and the text embeddings of class names. The second set measures cosine similarity between visual embeddings and centroids obtained through online clustering. These predictions are subsequently combined, considering their reliability as evaluated by Rényi Entropy (as outlined in Section~\ref{sec:renyi}), to yield the final results. For a comprehensive understanding of BaFTA's procedures, please also refer to Algorithm~\ref{alg:1} in Appendix~\ref{algorithmappendix}.

\section{Experiment and Results}
    \textbf{Baselines.} We conduct experiments in comparison of BaFTA with several benchmark models and algorithms. Our comparisons include the baseline model CLIP~\cite{radford2021learning}, the state-of-the-art test-time prompt-tuning algorithm TPT~\cite{shu2022tpt} and SwapPrompt~\cite{ma2024swapprompt}, and state-of-the-art training-free method CALIP~\cite{guo2023calip} with their officially reported scores. For CLIP, we report both single template (CLIP~\textit{single}), and multi-template version (CLIP~\textit{multi}). Furthermore, we have introduced CoOp~\cite{zhou2022coop} and PromptAlign~\cite{zhu2023prompt} as additional baseline models for comparison with few-shot prompt-tuning method, aligning with experiments from TPT.  

\textbf{Datasets.} We have conducted our experiments over two sets of datasets, following the experiment setup of \citet{shu2022tpt} and \citet{zhou2022coop}, which includes: \textbf{1)} ImageNet Robustness Evaluation with ImageNet~\cite{deng2009imagenet} and its Out-Of-Distribution (OOD) variants ImageNet-V2~\cite{imagenetv2}, ImageNet-R~\cite{imagenetr}, ImageNet-Sketch~\cite{imagenets} and ImageNet-A~\cite{imageneta}; \textbf{2)} Fine-Grained Datasets with Stanford Cars~\cite{cars}, Caltech101~\cite{caltech101}, Describable Textures (DTD)~\cite{dtd}, EuroSAT~\cite{helber2019eurosat}, FGVC Aircrafts~\cite{fgvc}, Food101~\cite{bossard2014food}, Flowers102~\cite{flower102}, Oxford-IIIT-Pets~\cite{pet}, UCF101~\cite{soomro2012ucf101} and SUN397~\cite{xiao2010sun}.
 

\textbf{Evaluation} We evaluate BaFTA under the test-time adaptation protocol following~\citet{shu2022tpt}, with  ViT-B/16 and ResNet50 checkpoints from CLIP as the baseline models for comparison and adaptation. In line with the TPT official implementation, we utilize a simple combination of \texttt{RandomResizedCrop} and \texttt{RandomFlip} to prepare 63 augmented views,  constituting a mini-batch of 64 images for each test image. For implementation details and hyper-parameters of BaFTA, please refer to Appendix~\ref{impmenetation}.

\label{experiment}
    \subsection{Main Results}

\begin{table}[!ht]
\centering
\resizebox{\textwidth}{!}{
\setlength{\tabcolsep}{0.5em}
\begin{tabular}{cc|ccccc} 
                & ImageNet       & ImageNet-A     & ImageNet-V2    & ImageNet-R     & ImageNet-Sketch & OOD Avg        \\ \toprule
CLIP~\textit{single} (ViT-B/16) & 66.73          & 47.87          & 60.86          & 73.98          & 46.09           & 57.20          \\ 
CLIP~\textit{multi} (ViT-B/16)  & 68.34          & 49.89          & 61.88          & 77.65          & 48.24           & 59.42          \\  \midrule
CoOp (16-shot)           & 71.51          & 49.71          & 64.20          & 75.21          & 47.99 & 59.28          \\
PromptAlign (16-shot) & - & 59.37 & 65.29 & 79.33 & 50.23 & 63.56 \\ 
TPT             & 68.98          & 54.77          & 63.45          & 77.06          & 47.94           & 60.81          \\
BaFTA           & \textbf{72.15} & \textbf{63.36} & \textbf{65.40} & \textbf{80.92} & \textbf{52.15}  & \textbf{65.46} \\ 
\bottomrule\toprule
CLIP~\textit{single} (RN50)     & 58.16          & 21.83          & 51.41          & 56.15          & 33.37           & 40.69          \\ 
CLIP~\textit{multi} (RN50)      & 59.81          & 23.24          & 52.91          & 60.72          & 35.48           & 43.09          \\ \midrule
CoOp (16-shot)           & 63.33          & 23.06          & 55.40          & 56.60          & 34.67            & 42.43          \\
TPT             & 60.74          & 26.67          & 54.70          & 59.11          & 35.09           & 43.89          \\
CALIP      & 60.57    & 23.96      & 53.70       & 60.81 & 35.61           & 43.52   \\
SwapPrompt & 61.41    & 24.42      & 52.93       & 60.25          & 38.13           & 43.93  \\
BaFTA           & \textbf{62.71} & \textbf{31.07} & \textbf{56.14} & \textbf{61.98}          & \textbf{38.22}  & \textbf{46.85}          \\ 
\bottomrule \\ 
\end{tabular}}
\caption{Comparison of top-1 accuracy on ImageNet and its Out-Of-Distribution (OOD) variants benchmarks. All methods evaluated in zero-shot classification setting, except that CoOp and PromptAlign are fine-tuned on ImageNet with 16 examples per category.}
\label{tab:imagenet} 
\end{table}

In Table~\ref{tab:imagenet} and Table~\ref{tab:fine-grained} we present the comprehensive results of backpropagation-free test-time algorithm BaFTA in comparison to baseline methods across five ImageNet robustness benchmarks and ten fine-grained classification benchmarks. 
As illustrated in Table~\ref{tab:imagenet} and Table~\ref{tab:fine-grained}, BaFTA significantly outperforms the baseline CLIP~\textit{multi} model, achieving improvements of 3.81\% on ImageNet, 6.04\% on OOD datasets, and 4.18\% on fine-grained datasets using the ViT-B/16 backbone. Comparable gains are observed with the ResNet-50 backbone as well.

When compared to the state-of-the-art test-time prompt tuning method TPT, BaFTA demonstrates substantial enhancements, recording a 3.17\% improvement on ImageNet, 4.65\% on OOD datasets, and 3.67\% on fine-grained datasets with the ViT-B/16 backbone. On the ResNet-50 backbone, we have also observed significant improvement of BaFTA over state-of-the-art methods TPT, SwapPrompt and CALIP on ImageNet, OOD variants and Fine-grained benchmarks. Remarkably, BaFTA achieves these results without relying on backpropagation training.


In comparison with few-shot fine-tuning methods such as CoOp and PromptAlign, BaFTA exhibits comparable performance on ImageNet and significantly better results on OOD and fine-grained datasets, with improvements of 5.55\% and 4.39\% over CoOp, and improvements of 1.9\% and 2.00\% over the state-of-the-art PromptAlign respectively, using the ViT-B/16 backbone. This indicates that unsupervised test-time adaptation algorithm BaFTA provides superior results compared to cross-domain generalization few-shot prompt tuning methods. 

The results presented in Tables~\ref{tab:imagenet} and~\ref{tab:fine-grained} have clearly demonstrated the effectiveness of BaFTA in improving the performance of vision-language base models at inference time. Notably, these enhancements are achieved without the need for backpropagation training or external resources, which solidifying BaFTA's position as a valuable and robust test-time adaptation method.

\begin{table}[!ht]
\centering
\resizebox{\textwidth}{!}{
\setlength{\tabcolsep}{0.5em}
\begin{tabular}{cccccccccccc} 
                                 & \rotatebox{90}{Average}        & \rotatebox{90}{Cars} & \rotatebox{90}{Caltech101} & \rotatebox{90}{DTD}  & \rotatebox{90}{EuroSAT} & \rotatebox{90}{FGVC} & \rotatebox{90}{Food101} & \rotatebox{90}{Flower102} & \rotatebox{90}{Pets} & \rotatebox{90}{UCF101} & \rotatebox{90}{SUN397}         \\ \midrule
CLIP~\textit{single} (ViT B/16)                      & 63.58          & 65.48          & 93.35          & 44.27          & 42.01          & 23.67          & 83.65          & 67.44          & 88.25          & 65.13          & 62.59          \\ 
CLIP~\textit{multi} (ViT B/16)                       & 64.59          & 66.11          & 93.55          & 45.04          & 50.42          & 23.22          & 82.86          & 66.99          & 86.92          & 65.16          & 65.63          \\ \midrule
CoOp (16-shot)                   & 63.88          & 64.51          & 93.70          & 41.92          & 46.39          & 18.47          & 85.30          & 68.71          & 89.14          & 66.55          & 64.15          \\
PromptAlign (16-shot) & 66.92   & 68.50 & 94.01      & 47.24 & 47.86   & 24.80 & 86.65   & 72.39     & 90.76 & 69.47  & 67.54  \\
TPT                              & 65.10          & 66.87          & \textbf{94.16} & 47.75          & 42.44          & 24.78          & 84.67 & 68.98          & 87.79          & 68.04          & 65.50          \\
BaFTA                            & \textbf{68.77} & \textbf{69.53} & 94.12          & \textbf{50.18} & \textbf{51.48} & \textbf{27.12} & \textbf{87.40} & \textbf{74.22} & \textbf{92.23} & \textbf{71.90} & \textbf{69.54} \\ \bottomrule\toprule
CLIP~\textit{single} (RN50)                      & 55.82          & 55.70          & 85.88          & 40.37          & 23.69          & 15.66          & 73.97          & 61.75          & 83.57          & 58.84          & 58.80          \\ 
CLIP~\textit{multi} (RN50)                       & 56.63          & 55.89          & 87.26          & 40.37          & 25.79          & 16.11          & 74.82          & 62.77          & 82.97          & 59.48          & 60.85          \\ \midrule
CoOp (16-shot)                   & 56.18          & 55.32          & 86.53          & 37.29          & 26.20          & 15.12          & 75.59          & 61.55          & 87.00          & 59.05          & 58.15          \\
TPT                              & 57.66          & 58.46 & 87.02          & 40.84          & 28.33 & 17.58          & 74.88          & 62.69          & 84.49          & 60.82          & 61.46          \\
 CALIP      & 59.35   & 56.27 & 87.71      & 42.39 & 38.90   & 17.76 & 77.42   & 66.38     & 86.31 & 61.72  & 58.59  \\
 SwapPrompt & -       & \textbf{58.88} & \textbf{89.90}      & -     & -       & -     & 75.08   & -         & 89.14 & -      & \textbf{63.93}  \\
BaFTA                            & \textbf{61.08} & 57.52          & 88.88 & \textbf{44.33} & \textbf{39.52} & \textbf{18.54} & \textbf{78.25} & \textbf{67.03} & \textbf{89.21} & \textbf{64.23} & 63.30 \\ \bottomrule \\
\end{tabular}}
\caption{Top-1 Accuracy on 10 Fine-grained Benchmarks. All baselines are evaluated in zero-shot classification setting, except CoOp and PromptAlign being fine-tuned on ImageNet with 16 examples per category. Note: SwapPrompt has provided results on DTD, EuroSAT, Flowers102 and UCF101 over different evaluation splits, which can not be directly compared. }
\label{tab:fine-grained}
\vspace{-0.5em}
\end{table}

\subsection{Ablation Studies}

In this section, we present the inference time efficiency analysis, the ablation studies on BaFTA components and the comparison of aggregation weight function.
Due to space constraint, please check Appendix~\ref{abla:alpha},~\ref{abla:beta},~\ref{abla:augviews} and~\ref{abla:projection} for the analysis over the Rényi Entropy Order $\alpha$, prediction balance weight $\beta$, number of augmentation views, and effectiveness of the projected embedding space.

\textbf{Inference Time Efficiency Analysis.}
\label{abla:time}
We conducted experiments to demonstrate the efficiency difference between TPT and BaFTA using both ViT-B/16 and RN50 backbones. While TPT requires an average of 873 and 841 millisecond to complete the inference on a single image with the ViT-B/16 and RN50, BaFTA completes such inference with only 158.7 and 183.8 millisecond over the corresponding backbones. The inference time per example (in milliseconds) was calculated by recording the total time required to complete the inference on 10,000 ImageNet examples. All experiments were conducted with a single NVIDIA A40 GPU.
As indicated by the results, 
BaFTA exhibits a notable advantage, being approximately 5 times faster than TPT. The significant difference in inference time for TPT can be attributed to two main factors: 1) TPT requires two forward passes and one backward pass in each iteration, whereas BaFTA requires only a single forward pass; 2) TPT requires recomputation of classification embeddings through the text encoder at each forward pass, while BaFTA conducts the text encoder once offline and updates the classification embeddings directly in the embedding space during inference.

\textbf{Component Effectiveness Analysis.}
\label{abla:component}
In Table~\ref{component}, we assess the impact of various BaFTA components on its overall performance. Our evaluation focuses on: 1) the utilization of multi-template (MT) prompts provided by CLIP~\cite{radford2021learning}; 2) the choices of different classification weights, including regular text embeddings (TE), online clustering centroids (OC), or a combination of both (TE \& OC); and 3) the choices of various augmented view aggregation strategies, such as Rényi Entropy aggregation, simple averaging, or averaging predictions with top 10\% confidence.

The inherent computational complexity of prompt tuning method TPT limits it to single-template base models. Additionally, despite utilizing augmentations for prompt tuning, TPT only employs original image for its predictions.
To ensure a fair comparison and demonstrate the effectiveness of BaFTA innovations such as Rényi Entropy aggregation and projected online clustering, we evaluated BaFTA~\textit{single} and TPT-Agg.
BaFTA~\textit{single} utilizes a single-template CLIP as its base model, while TPT-Agg aggregates predictions from top 10\% high-confidence augmented views. Experiment results reveal that BaFTA~\textit{single} outperforms TPT-Agg by 1.61\%, demonstrating the effectiveness of BaFTA's innovations. 
It is worth noting that BaFTA results can be further improved with the multi-template prompts, and it achieves these superior results via a backprop-free approach which is about 5 times faster than TPT. 

Results for BaFTA-TE and BaFTA-OC, as summarized in Table~\ref{component}, correspond to aggregated predictions $p_i, \hat{p_i}$ detailed in Equation~\ref{eq:renyi2}. 
With the help from Rényi Entropy aggregation, both BaFTA-TE and BaFTA-OC improve the baseline model, achieving 3.24\% improvement with the text embedding predictions, and 0.83\% improvement with the online clustering predictions. 
As discussed in Section~\ref{sec:renyi}, the smaller improvement with BaFTA-OC is attributed to potential biases and early-stage instability in clustering. 
However, combining TE and OC predictions significantly enhances BaFTA's performance, yielding an average improvement of 4.65\% across 15 datasets over the baseline CLIP model.
This indicates the effectiveness of combining both text embedding predictions and online clustering predictions. 

Lastly, to validate the contribution of Rényi Entropy aggregation against naive simple average, we assessed BaFTA-Avg, wherein Rényi Entropy weighted averaging was substituted with simple averaging. This comparison revealed that BaFTA surpasses BaFTA-Avg by 1.87\%, clearly demonstrating the effectiveness of Rényi Entropy aggregation.


\begin{table}[!ht]
\centering
\begin{tabular}{ccccc}
                            & Multi-Template                & Classification Weight                & Prediction Aggregation          & Accuracy     \\ \toprule
CLIP~\textit{single}        & \xmark            & TE                         & None                     & 62.09   \\ 
TPT                         & \xmark                 & TE                         & None                     & 64.21   \\
TPT-Agg                     & \xmark                 & TE                         & Top 10\% Avg.          & 64.34   \\
TPT-Rényi                     & \xmark                 & TE                         & Rényi Entropy          & 64.42   \\

BaFTA~\textit{single}       & \xmark            & TE \& OC                    & Rényi Entropy         & 65.95   \\ \midrule
CLIP~\textit{multi}         & \cmark            & TE                         & None                     & 63.46   \\ 
BaFTA-TE                    & \cmark            & TE                    & Rényi Entropy                & 66.70   \\
BaFTA-OC                    & \cmark            & OC                    & Rényi Entropy                & 64.29   \\
BaFTA-Avg                   & \cmark            & TE \& OC                    & Average                & 66.24   \\
BaFTA                       & \cmark            & TE \& OC                    & Rényi Entropy                & \textbf{68.11}   \\ \bottomrule \\
\end{tabular}
\caption{Effectiveness of BaFTA components. All results are averaged accuracies over 15 datasets, produced with CLIP (ViT-B/16). TE: Text Embeddings of Category Names, OC: Online Clustering Centroids. 
Please refer to Appendix~\ref{fullresult} for complete results. }
\label{component}
\end{table}

\begin{table}[!ht]
    \centering
    \setlength{\tabcolsep}{0.5em}
    \resizebox{\linewidth}{!}{
    \begin{tabular}{cccccccc} 
       $\weight(p^b)$     & 1 & $\max(p^b)$ & $\mathbbm{1}[H(p^b)<\tau]$ & $\hat{H}(p^b)$ & $Re_{0.25}(p^b)$ & $Re_{0.5}(p^b)$      & $Re_{0.75}(p^b)$ \\ \toprule
    Accuracy & 69.43   & 70.60   & 70.34       & 69.65   & 70.85      & \textbf{71.00} & 70.97  \\ \bottomrule \\ 
    \end{tabular}}
    \caption{Comparison on weighting function $\weight(p^b)$ that merges text embedding predictions $p_b$ from augmented views into aggregated prediction of $p=\sum_b \weight(p^b)p^b$. All results are top-1 accuracy reported with CLIP (ViT-B/16) on ImageNet with 64 augmented views per test example.} 
    \label{tab:vote} 
\end{table}

\textbf{Aggregation Function Comparison.}
\label{abla:renyi}
In Table~\ref{tab:vote}, we present the ablation results on choice of aggregation function that merges text embedding predictions results from augmented views. From left to right, we have: 1) $\weight(p^b)=1$: simple averaging; 2) $\max(p^b)$: weighted-average prediction with confidence estimated by maximum entry of $p^b$; 3) $\mathbbm{1}[H(p^b)>\tau]$: average of low-entropy (high-confidence) predictions, as adopted by TPT; 4) $\hat{H}(p^b)$: weighted-average prediction with confidence estimated by the normalized entropy $\hat{H}(p^b)={(H^{max}-H(p^b))}/{H^{max}}$; 5) $Re_{\alpha}(p^b)$: weighted-average prediction with confidence estimated by Rényi entropy $Re(p^b)$, with entropy order $\alpha=0.25,0.50,0.75$. As shown in Table~\ref{tab:vote}, Rényi Entropy at order of 0.50 provides the best results over all the other options. Please also refer to Appendix~\ref{abla:alpha} for more analysis on Rényi Entropy Order.

\section{Conclusion}
    In this work, we have focused on enhancing the performance of large-scale pre-trained vision-language models at inference time, in the context of zero-shot image classification. We have introduced a novel backpropagation-free test-time adaptation algorithm BaFTA. Unlike the previous methods which fine-tune text prompts to refine class embeddings, our approach directly estimates class centroids by performing online clustering within a projected embedding space that aligns text and visual embeddings. We have also proposed a dynamic aggregation method, which leverages predictions from both estimated and original class embeddings, as well as distinct augmented views, by assessing their reliability with Rényi Entropy.
Our comprehensive experiments have shown that BaFTA consistently outperforms existing state-of-the-art test-time adaptation methods, achieving significant performance improvements at a considerably faster speed. BaFTA has demonstrated the feasibility of a stable and efficient backpropagation-free solution achieving strong performance in VLM adaptation, and offers a viable solution for practical real-world applications.

\newpage
\bibliography{example_paper}

\begin{thebibliography}{36}
\providecommand{\natexlab}[1]{#1}
\providecommand{\url}[1]{\texttt{#1}}
\expandafter\ifx\csname urlstyle\endcsname\relax
  \providecommand{\doi}[1]{doi: #1}\else
  \providecommand{\doi}{doi: \begingroup \urlstyle{rm}\Url}\fi

\bibitem[Barbakh \& Fyfe(2008)Barbakh and Fyfe]{barbakh2008online}
Barbakh, W. and Fyfe, C.
\newblock Online clustering algorithms.
\newblock \emph{International journal of neural systems}, 18\penalty0 (03):\penalty0 185--194, 2008.

\bibitem[Bossard et~al.(2014)Bossard, Guillaumin, and Gool]{bossard2014food}
Bossard, L., Guillaumin, M., and Gool, L.~V.
\newblock Food-101--mining discriminative components with random forests.
\newblock In \emph{European conference on computer vision}, pp.\  446--461. Springer, 2014.

\bibitem[Cimpoi et~al.(2014)Cimpoi, Maji, Kokkinos, Mohamed, , and Vedaldi]{dtd}
Cimpoi, M., Maji, S., Kokkinos, I., Mohamed, S., , and Vedaldi, A.
\newblock Describing textures in the wild.
\newblock In \emph{Proceedings of the {IEEE} Conf. on Computer Vision and Pattern Recognition ({CVPR})}, 2014.

\bibitem[Deng et~al.(2009)Deng, Dong, Socher, Li, Li, and Fei-Fei]{deng2009imagenet}
Deng, J., Dong, W., Socher, R., Li, L.-J., Li, K., and Fei-Fei, L.
\newblock Imagenet: A large-scale hierarchical image database.
\newblock In \emph{2009 IEEE conference on computer vision and pattern recognition}, pp.\  248--255. Ieee, 2009.

\bibitem[Ge et~al.(2023)Ge, Ren, Gallagher, Wang, Yang, Adam, Itti, Lakshminarayanan, and Zhao]{ge2023improving}
Ge, Y., Ren, J., Gallagher, A., Wang, Y., Yang, M.-H., Adam, H., Itti, L., Lakshminarayanan, B., and Zhao, J.
\newblock Improving zero-shot generalization and robustness of multi-modal models.
\newblock In \emph{Proceedings of the IEEE/CVF Conference on Computer Vision and Pattern Recognition}, pp.\  11093--11101, 2023.

\bibitem[Guo et~al.(2023)Guo, Zhang, Qiu, Ma, Miao, He, and Cui]{guo2023calip}
Guo, Z., Zhang, R., Qiu, L., Ma, X., Miao, X., He, X., and Cui, B.
\newblock Calip: Zero-shot enhancement of clip with parameter-free attention.
\newblock In \emph{Proceedings of the AAAI Conference on Artificial Intelligence}, volume~37, pp.\  746--754, 2023.

\bibitem[Helber et~al.(2019)Helber, Bischke, Dengel, and Borth]{helber2019eurosat}
Helber, P., Bischke, B., Dengel, A., and Borth, D.
\newblock Eurosat: A novel dataset and deep learning benchmark for land use and land cover classification.
\newblock \emph{IEEE Journal of Selected Topics in Applied Earth Observations and Remote Sensing}, 12\penalty0 (7):\penalty0 2217--2226, 2019.

\bibitem[Hendrycks et~al.(2021{\natexlab{a}})Hendrycks, Basart, Mu, Kadavath, Wang, Dorundo, Desai, Zhu, Parajuli, Guo, Song, Steinhardt, and Gilmer]{imagenetr}
Hendrycks, D., Basart, S., Mu, N., Kadavath, S., Wang, F., Dorundo, E., Desai, R., Zhu, T., Parajuli, S., Guo, M., Song, D., Steinhardt, J., and Gilmer, J.
\newblock The many faces of robustness: A critical analysis of out-of-distribution generalization.
\newblock \emph{ICCV}, 2021{\natexlab{a}}.

\bibitem[Hendrycks et~al.(2021{\natexlab{b}})Hendrycks, Zhao, Basart, Steinhardt, and Song]{imageneta}
Hendrycks, D., Zhao, K., Basart, S., Steinhardt, J., and Song, D.
\newblock Natural adversarial examples.
\newblock In \emph{Proceedings of the IEEE/CVF Conference on Computer Vision and Pattern Recognition}, pp.\  15262--15271, 2021{\natexlab{b}}.

\bibitem[Hu et~al.(2023)Hu, Zhang, Xia, Chen, Luo, Sun, Wang, Qiao, Zeng, Sun, Kuo, and Nevatia]{xuefeng2023reclip}
Hu, X., Zhang, K., Xia, L., Chen, A., Luo, J., Sun, Y., Wang, K., Qiao, N., Zeng, X., Sun, M., Kuo, C.-H., and Nevatia, R.
\newblock Reclip: Refine contrastive language image pre-training with source free domain adaptation.
\newblock \emph{arXiv preprint arXiv:2308.03793}, 2023.

\bibitem[Jia et~al.(2021)Jia, Yang, Xia, Chen, Parekh, Pham, Le, Sung, Li, and Duerig]{jia2021scaling}
Jia, C., Yang, Y., Xia, Y., Chen, Y.-T., Parekh, Z., Pham, H., Le, Q., Sung, Y.-H., Li, Z., and Duerig, T.
\newblock Scaling up visual and vision-language representation learning with noisy text supervision.
\newblock In \emph{International Conference on Machine Learning}, pp.\  4904--4916. PMLR, 2021.

\bibitem[Karmanov et~al.(2024)Karmanov, Guan, Lu, Saddik, and Xing]{karmanov2024efficient}
Karmanov, A., Guan, D., Lu, S., Saddik, A.~E., and Xing, E.
\newblock Efficient test-time adaptation of vision-language models.
\newblock \emph{arXiv preprint arXiv:2403.18293}, 2024.

\bibitem[Krause et~al.(2013)Krause, Stark, Deng, and Fei-Fei]{cars}
Krause, J., Stark, M., Deng, J., and Fei-Fei, L.
\newblock 3d object representations for fine-grained categorization.
\newblock In \emph{4th International IEEE Workshop on 3D Representation and Recognition (3dRR-13)}, Sydney, Australia, 2013.

\bibitem[Laptev \& Ginsburg(2023)Laptev and Ginsburg]{laptev2023fast}
Laptev, A. and Ginsburg, B.
\newblock Fast entropy-based methods of word-level confidence estimation for end-to-end automatic speech recognition.
\newblock In \emph{2022 IEEE Spoken Language Technology Workshop (SLT)}, pp.\  152--159. IEEE, 2023.

\bibitem[Li et~al.(2022)Li, Andreeto, Ranzato, and Perona]{caltech101}
Li, Andreeto, Ranzato, and Perona.
\newblock Caltech 101, Apr 2022.

\bibitem[Liang et~al.(2023)Liang, He, and Tan]{liang2023comprehensive}
Liang, J., He, R., and Tan, T.
\newblock A comprehensive survey on test-time adaptation under distribution shifts.
\newblock \emph{arXiv preprint arXiv:2303.15361}, 2023.

\bibitem[Liang et~al.(2022)Liang, Zhang, Kwon, Yeung, and Zou]{liang2022mind}
Liang, V.~W., Zhang, Y., Kwon, Y., Yeung, S., and Zou, J.~Y.
\newblock Mind the gap: Understanding the modality gap in multi-modal contrastive representation learning.
\newblock \emph{Advances in Neural Information Processing Systems}, 35:\penalty0 17612--17625, 2022.

\bibitem[Ma et~al.(2024)Ma, Zhang, Guo, and Xu]{ma2024swapprompt}
Ma, X., Zhang, J., Guo, S., and Xu, W.
\newblock Swapprompt: Test-time prompt adaptation for vision-language models.
\newblock \emph{Advances in Neural Information Processing Systems}, 36, 2024.

\bibitem[Maji et~al.(2013)Maji, Kannala, Rahtu, Blaschko, and Vedaldi]{fgvc}
Maji, S., Kannala, J., Rahtu, E., Blaschko, M., and Vedaldi, A.
\newblock Fine-grained visual classification of aircraft.
\newblock Technical report, 2013.

\bibitem[Manli et~al.(2022)Manli, Weili, De-An, Zhiding, Tom, Anima, and Chaowei]{shu2022tpt}
Manli, S., Weili, N., De-An, H., Zhiding, Y., Tom, G., Anima, A., and Chaowei, X.
\newblock Test-time prompt tuning for zero-shot generalization in vision-language models.
\newblock In \emph{NeurIPS}, 2022.

\bibitem[Nilsback \& Zisserman(2008)Nilsback and Zisserman]{flower102}
Nilsback, M.-E. and Zisserman, A.
\newblock Automated flower classification over a large number of classes.
\newblock In \emph{2008 Sixth Indian Conference on Computer Vision, Graphics \& Image Processing}, pp.\  722--729. IEEE, 2008.

\bibitem[Niu et~al.(2023)Niu, Wu, Zhang, Wen, Chen, Zhao, and Tan]{niu2023towards}
Niu, S., Wu, J., Zhang, Y., Wen, Z., Chen, Y., Zhao, P., and Tan, M.
\newblock Towards stable test-time adaptation in dynamic wild world.
\newblock \emph{arXiv preprint arXiv:2302.12400}, 2023.

\bibitem[Park \& D’Amico(2023)Park and D’Amico]{park2023robust}
Park, T.~H. and D’Amico, S.
\newblock Robust multi-task learning and online refinement for spacecraft pose estimation across domain gap.
\newblock \emph{Advances in Space Research}, 2023.

\bibitem[Parkhi et~al.(2012)Parkhi, Vedaldi, Zisserman, and Jawahar]{pet}
Parkhi, O.~M., Vedaldi, A., Zisserman, A., and Jawahar, C.
\newblock Cats and dogs.
\newblock In \emph{2012 IEEE conference on computer vision and pattern recognition}, pp.\  3498--3505. IEEE, 2012.

\bibitem[Radford et~al.(2021)Radford, Kim, Hallacy, Ramesh, Goh, Agarwal, Sastry, Askell, Mishkin, Clark, et~al.]{radford2021learning}
Radford, A., Kim, J.~W., Hallacy, C., Ramesh, A., Goh, G., Agarwal, S., Sastry, G., Askell, A., Mishkin, P., Clark, J., et~al.
\newblock Learning transferable visual models from natural language supervision.
\newblock In \emph{International Conference on Machine Learning}, pp.\  8748--8763. PMLR, 2021.

\bibitem[Recht et~al.(2019)Recht, Roelofs, Schmidt, and Shankar]{imagenetv2}
Recht, B., Roelofs, R., Schmidt, L., and Shankar, V.
\newblock Do imagenet classifiers generalize to imagenet?
\newblock In \emph{International conference on machine learning}, pp.\  5389--5400. PMLR, 2019.

\bibitem[Samadh et~al.(2023)Samadh, Gani, Hussein, Khattak, Naseer, Khan, and Khan]{samadh2023align}
Samadh, J. H.~A., Gani, H., Hussein, N.~H., Khattak, M.~U., Naseer, M., Khan, F., and Khan, S.
\newblock Align your prompts: Test-time prompting with distribution alignment for zero-shot generalization.
\newblock In \emph{Thirty-seventh Conference on Neural Information Processing Systems}, 2023.

\bibitem[Soomro et~al.(2012)Soomro, Zamir, and Shah]{soomro2012ucf101}
Soomro, K., Zamir, A.~R., and Shah, M.
\newblock Ucf101: A dataset of 101 human actions classes from videos in the wild.
\newblock \emph{arXiv preprint arXiv:1212.0402}, 2012.

\bibitem[Tanwisuth et~al.(2023)Tanwisuth, Zhang, Zheng, He, and Zhou]{tanwisuth2023pouf}
Tanwisuth, K., Zhang, S., Zheng, H., He, P., and Zhou, M.
\newblock Pouf: Prompt-oriented unsupervised fine-tuning for large pre-trained models.
\newblock In \emph{International Conference on Machine Learning}, 2023.

\bibitem[Udandarao et~al.(2022)Udandarao, Gupta, and Albanie]{udandarao2022sus}
Udandarao, V., Gupta, A., and Albanie, S.
\newblock Sus-x: Training-free name-only transfer of vision-language models.
\newblock \emph{arXiv preprint arXiv:2211.16198}, 2022.

\bibitem[Wang et~al.(2020)Wang, Shelhamer, Liu, Olshausen, and Darrell]{wang2020tent}
Wang, D., Shelhamer, E., Liu, S., Olshausen, B., and Darrell, T.
\newblock Tent: Fully test-time adaptation by entropy minimization.
\newblock \emph{arXiv preprint arXiv:2006.10726}, 2020.

\bibitem[Wang et~al.(2019)Wang, Ge, Lipton, and Xing]{imagenets}
Wang, H., Ge, S., Lipton, Z., and Xing, E.~P.
\newblock Learning robust global representations by penalizing local predictive power.
\newblock In \emph{Advances in Neural Information Processing Systems}, pp.\  10506--10518, 2019.

\bibitem[Xiao et~al.(2010)Xiao, Hays, Ehinger, Oliva, and Torralba]{xiao2010sun}
Xiao, J., Hays, J., Ehinger, K.~A., Oliva, A., and Torralba, A.
\newblock Sun database: Large-scale scene recognition from abbey to zoo.
\newblock In \emph{2010 IEEE computer society conference on computer vision and pattern recognition}, pp.\  3485--3492. IEEE, 2010.

\bibitem[Zhou et~al.(2022{\natexlab{a}})Zhou, Yang, Loy, and Liu]{zhou2022cocoop}
Zhou, K., Yang, J., Loy, C.~C., and Liu, Z.
\newblock Conditional prompt learning for vision-language models.
\newblock In \emph{IEEE/CVF Conference on Computer Vision and Pattern Recognition (CVPR)}, 2022{\natexlab{a}}.

\bibitem[Zhou et~al.(2022{\natexlab{b}})Zhou, Yang, Loy, and Liu]{zhou2022coop}
Zhou, K., Yang, J., Loy, C.~C., and Liu, Z.
\newblock Learning to prompt for vision-language models.
\newblock \emph{International Journal of Computer Vision (IJCV)}, 2022{\natexlab{b}}.

\bibitem[Zhu et~al.(2023)Zhu, Niu, Han, Wu, and Zhang]{zhu2023prompt}
Zhu, B., Niu, Y., Han, Y., Wu, Y., and Zhang, H.
\newblock Prompt-aligned gradient for prompt tuning.
\newblock In \emph{Proceedings of the IEEE/CVF International Conference on Computer Vision}, pp.\  15659--15669, 2023.

\end{thebibliography}
\bibliographystyle{icml2024}

\newpage
\appendix


\section{Full Algorithm of BaFTA}
\label{algorithmappendix}

In Algorithm \ref{alg:1} we present the detailed procedures in BaFTA. 

\begin{algorithm}[h]
    \caption{BaFTA: Backprop-Free Test-Time Adaptation for Zero-Shot Vision-Language Models.}
    \label{alg:1}
    \begin{algorithmic}
    \REQUIRE Vision Language Pre-trained Model $M=\{M_v,M_t\}$
    \REQUIRE Test Samples $X=\{x_i\}_{i=1}^I$
    \REQUIRE Class Names $C=\{c_j\}_{j=1}^J$
    \REQUIRE Template Prompts $\{\theta_z\}_{z=1}^Z$

    \STATE $ t_j \gets \frac{1}{Z}\sum_{z} M_t(\theta_z(c_j)$ \COMMENT{Prepare multi-templates text embeddings for each class}
    \STATE $\hat{t_j} \gets P^*(t_j|\{t_1,...,t_J\})$ \COMMENT{Projected text embeddings (Eq~\ref{proj_eq})} 
    \STATE $w_j \gets \hat{t_j}, k_j\gets 0$ \COMMENT{Initialize class centroids $w_j$ and counter $k_j$ for each class}
    \FOR{i $\gets 1$ to $I$} 
        \STATE $\{x_i^b\}_{b=1}^B \gets A(x_i)$ \COMMENT{Generate $B$ views with random augmentation function $A(\cdot)$}
        \STATE $v_i^b \gets M_v(x_i^b)$
        \COMMENT{Visual embedding for each augmented views}
        \STATE $\hat{v_i^b} \gets P^*(v_i^b)$
        \COMMENT{Projected visual embedding (Eq~\ref{proj_eq})}
        \STATE $ p_{i}^b \gets softmax\left(\left[cos(v_i^b, t_j)\right]_{j=1}^J\right)$ \\\COMMENT{Cosine-similarity between visual embedding $v_i^b$ and text embedding $t_j^b$, (Eq~\ref{pd1})}
        \STATE $ \hat{p}_{i}^b \gets softmax\left(\left[cos(\hat{v_i^b}, w_j)\right]_{j=1}^J\right)$ \\\COMMENT{Cosine-similarity between projected visual embedding $\hat{v_i^b}$ and class centroids $w_j^b$, (Eq~\ref{pd2})}
        \STATE $\Tilde{p_i}\gets \frac{\beta}{R_1}\sum_{b} Re(p_{i}^b)p_{i}^b + \frac{1}{R_2}\sum_{b} Re(\hat{p}_{i}^b)\hat{p}_{i}^b$ \COMMENT{Prediction Aggregation (Eq. \ref{eq:renyi2})}
        \STATE $y_i \gets \argmax_j \Tilde{p_i}$ \COMMENT{Get prediction for example $x_i$}
        \STATE $\hat{v_i} \gets \frac{1}{B}\sum_{b=1}^B \hat{v_i^b}$
        \STATE $w_j \gets \left(k_{j}w_j + \hat{v_i}\right)/\lVert\left(k_{j}w_j + \hat{v_i}\right)\rVert$, $k_j\gets k_j+1$ for $j=y_i$ \\
        \COMMENT{Updates centroids and counter on predicted class $y_i$ (Eq \ref{online_cluster_eq})}
        \STATE Output $y_i$ as prediction for $x_i$
    \ENDFOR
    \end{algorithmic}
\end{algorithm}

\section{Full Results of Ablation Study on Effectiveness of BaFTA Components}
\label{fullresult}

In Table~\ref{tab:abla1full} we present the complete results on the ablation analysis of BaFTA components as mentioned in Table~\ref{abla:component}.

\begin{table}[!h]
\resizebox{\textwidth}{!}{
\setlength{\tabcolsep}{0.3em}
\begin{tabular}{ccccccccccccccccc}
  & \rotatebox{90}{Average}        & \rotatebox{90}{ImageNet}       & \rotatebox{90}{ImageNet-A}     & \rotatebox{90}{ImageNet-V2}    & \rotatebox{90}{ImageNet-R}     & \rotatebox{90}{ImageNet-S}         & \rotatebox{90}{Cars} & \rotatebox{90}{Caltech101} & \rotatebox{90}{DTD}  & \rotatebox{90}{EuroSAT} & \rotatebox{90}{FGVC} & \rotatebox{90}{Food101} & \rotatebox{90}{Flower102} & \rotatebox{90}{Pets} & \rotatebox{90}{UCF101} & \rotatebox{90}{SUN397}         \\ \toprule
CLIP~\textit{single}              & 62.09   & 66.73                         & 47.87      & 60.86       & 73.98                         & 46.09                         & 65.48 & 93.35      & 44.27 & 42.01   & 23.67 & 83.65                         & 67.44     & 88.25 & 65.13  & 62.59  \\ \midrule
TPT                      & 64.21   & 68.98    & 54.77      & 63.45       & 77.06      & 47.94           & 66.87 & 94.16      & 47.75 & 42.44   & 24.78 & 84.67   & 68.98     & 87.79 & 68.04  & 65.50  \\
TPT-Agg             & 64.34   & 69.00    & \textbf{61.54}      & 64.22       & 77.46      & 48.10           & 67.69 & 93.26      & 45.45 & 40.83   & 23.49 & 84.48   & 68.98     & 87.03 & 68.83  & 64.77  \\ 
TPT-Rényi & 64.42   & 69.50 & 60.61      & \textbf{64.29} & 77.23   & 48.21 & \textbf{68.44}   & 93.31     & 45.74 & 39.65  & 23.82  & 84.77 & 69.35 & 87.16 & \textbf{68.99} & 65.22 \\
BaFTA~\textit{single}             & \textbf{65.95}   & \textbf{70.50}                & 61.28      & 64.25       & \textbf{78.19}      & \textbf{49.24}           & 68.23 & \textbf{93.79}      & \textbf{48.52} & \textbf{45.90}   & \textbf{25.23} & \textbf{87.28}   & \textbf{70.28}     & \textbf{90.41} & 68.15  & \textbf{68.05}  \\ \midrule\midrule
CLIP~\textit{multi}               & 63.46   & 68.34    & 49.89      & 61.88       & 77.65      & 48.24           & 66.11 & 93.55      & 45.04 & 50.42   & 23.22 & 82.86   & 66.99     & 86.92 & 65.16  & 65.63  \\ \midrule
BaFTA-Avg                & 66.24   & 70.23    & 56.25      & 65.35       & 75.29      & 49.00           & 68.47 & 93.47      & 49.53 & 47.79   & 26.37 & 86.05   & 74.14     & 92.15 & 70.87  & 68.61  \\
BaFTA-TE                 & 66.70   & 71.00    & 62.85      & 65.39       & 80.70      & 50.81           & 67.67 & 94.00      & 47.04 & 48.96   & \textbf{27.45} & 86.94   & 71.54     & 89.34 & 69.79  & 66.97  \\
BaFTA-OC                 & 64.29   & 67.34    & 55.28      & 50.21       & 76.81      & 48.00           & 64.61 & 92.21      & 49.17 & 49.81   & 25.32 & 85.86   & 71.21     & 92.18 & 69.63  & 66.74  \\
BaFTA                    & \textbf{68.11}   & \textbf{72.15}    & \textbf{63.36}      & \textbf{65.40}       & \textbf{80.92}      & \textbf{52.15}           & \textbf{69.53} & \textbf{94.12}      & \textbf{50.18} & \textbf{51.48}   & 27.12 & \textbf{87.40}   & \textbf{74.22}     & \textbf{92.23} & \textbf{71.90}  & \textbf{69.54}  \\ \bottomrule \\
\end{tabular}}
\caption{Ablation studies on the effectiveness of BaFTA components, as described in Table~\ref{component} and discussed in Section~\ref{abla:component}. All results produced with CLIP (ViT-B/16).}
\label{tab:abla1full}
\end{table}

\section{Orthogonal Improvement over Few-Shot Fine-tuned Prompts}
\begin{table}[!ht]
\centering
\resizebox{\textwidth}{!}{
\setlength{\tabcolsep}{0.5em}
\begin{tabular}{cc|ccccc} 
                & ImageNet       & ImageNet-A     & ImageNet-V2    & ImageNet-R     & ImageNet-Sketch & OOD Avg        \\ \toprule
CLIP~\textit{single} (ViT-B/16) & 66.73          & 47.87          & 60.86          & 73.98          & 46.09           & 57.20          \\ 
CLIP~\textit{multi} (ViT-B/16)  & 68.34          & 49.89          & 61.88          & 77.65          & 48.24           & 59.42          \\  \midrule
CoOp (16-shot)           & 71.51          & 49.71          & 64.20          & 75.21          & 47.99 & 59.28          \\
TPT+CoOp                 & 73.61          & 57.95          & 66.83          & 77.27          & 49.29 & 62.84          \\
BaFTA + CoOp    & \textbf{75.10} & \textbf{63.84} & \textbf{67.55} & \textbf{81.00} & \textbf{52.46}  & \textbf{66.21}      \\ 
\bottomrule\toprule
CLIP~\textit{single} (RN50)     & 58.16          & 21.83          & 51.41          & 56.15          & 33.37           & 40.69          \\ 
CLIP~\textit{multi} (RN50)      & 59.81          & 23.24          & 52.91          & 60.72          & 35.48           & 43.09          \\ \midrule
CoOp (16-shot)           & 63.33          & 23.06          & 55.40          & 56.60          & 34.67            & 42.43          \\
TPT+CoOp                 & 64.73          & 30.32 & 57.83          & 58.99          & 35.86            & 45.75          \\
BaFTA + CoOp      & \textbf{66.56} & \textbf{33.49} & \textbf{58.84} & \textbf{62.01} & \textbf{38.46}  & \textbf{48.20}             \\ 
\bottomrule \\ 
\end{tabular}}
\caption{Orthogonal Improvements over Few-Shot Fine-tuned Prompts. All methods evaluated in zero-shot classification setting.}
\label{tab:imagenet} 
\end{table}

Following~\citet{shu2022tpt}, we provide comparison of BaFTA and TPT over ImageNet fine-tuned prompts CoOp. As shown by the results, both TPT and BaFTA can be combined with fine-tuned prompts and provides orthogonal enhancements. BaFTA+CoOp achieves improvements of 6.86\% and 6.79\% over the baseline CLIP ViT-B/16 model on ImageNet and OOD datasets, surpassing TPT, CoOp, and TPT+CoOp by significant margins.

\section{Implementation Details} 
\label{impmenetation}

Unless otherwise specified, we use the same hyper-parameters for all BaFTA experiments, with $\beta=2$ and $\alpha=0.5$. We have employed the exponential form of Renyi Entropy following~\citet{laptev2023fast}. All BaFTA results are reported with a warm-up schedule of $10J$ examples ($J$ as number of class) before the online clustering predictions aggregated into final prediction. For the embedding projection matrix, we use $U'=[e_2,...,e_J]$ for all datasets, except for datasets with more than 150 categories such as ImageNet, we use $U'=[e_2,...,e_{150}]$ for best performance. 

For experiments on-top-of the CoOp, we use the 16-shot fine-tuned model and ensemble the predictions generated from CoOp embeddings with our predictions using Rényi entropy. Instead of directly replacing the prompts, we adopt this approach because we have observed that CoOp embeddings sometimes perform less effectively than the multi-template embeddings provided by CLIP. For all other BaFTA results, we use official template sets provided by CLIP to generate the text embeddings. 

For experiment on time efficiency analysis, all experiments were conducted on a computation node equipped with an AMD EPYC 7313 CPU (32 cores), 256 GB memory, and a single NVIDIA A40 GPU (48GB).

\section{Ablation Studies on Rényi Entropy Order $\alpha$. }
\label{abla:alpha}
\begin{figure*}[!ht]
    \centering
    \includegraphics[width=\linewidth]{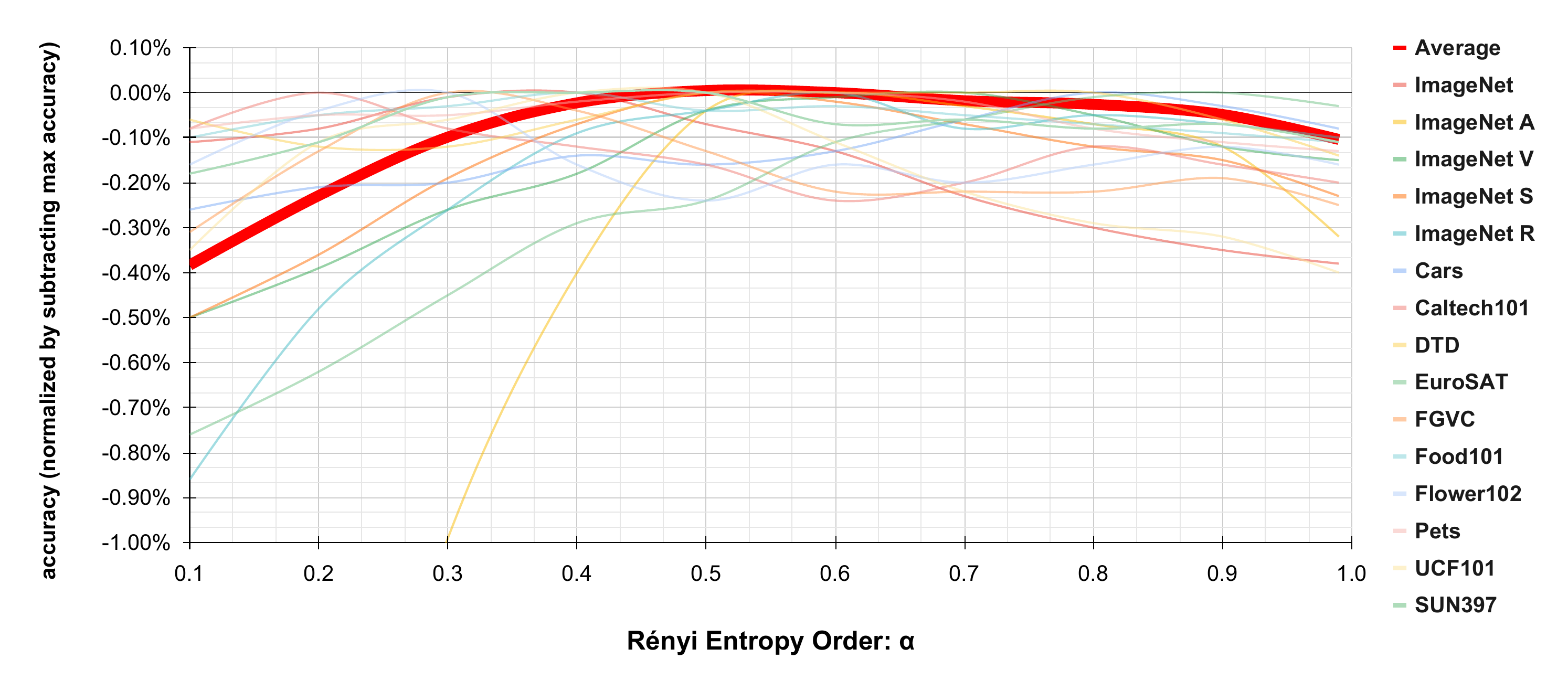}
    \caption{$\alpha$-accuracy curves on 15 datasets, with $\alpha\in[0.1,0.99]$. In order to fit all curves into one plot with unified value range, all curves are normalized by subtracting the maximum accuracy within the curve. The bold red curve represents the averaged accuracy over 15 datasets, achieves its maximum value at $\alpha=0.5$ and $\alpha=0.6$. This plot indicates that prediction aggregation accuracy is not highly sensitive to the choice of $\alpha$, with most curves exhibiting less than a 0.3\% change in accuracy across the $\alpha$ range [0.1, 0.99]}
    \label{fig:alpha}
\end{figure*}

In order to assess the sensitivity of Rényi Entropy aggregation performance to the entropy order $\alpha$, we analyze the accuracy of Rényi Entropy aggregated predictions from augmented views with varying $\alpha$, specifically $\alpha\in\{0.1, 0.2, 0.3,0.4,0.5,0.6,0.7,0.8,0.9,0.99\}$, over all 15 datasets used in our study. 
In all experiments, we employ CLIP-ViT-B/16 as the base model and evaluate BaFTA-RA to investigate the influence of $\alpha$ on prediction aggregation without the impact of online clustering results.

Figure~\ref{fig:alpha} illustrates the $\alpha-$accuracy curve across all 15 datasets. The curves are normalized by subtracting the maximum value within each curve, ensuring they are plotted within the same value range. The bold red curves represent the averaged accuracy over the 15 datasets, revealing that the average performance peaks at $\alpha=0.5$ and $\alpha=0.6$. Additionally, the plot indicates that prediction aggregation accuracy is relatively insensitive to the choice of $\alpha$, with most curves exhibiting less than a 0.3\% change in accuracy across the $\alpha$ range [0.1, 0.99]. Most datasets achieve peak performance with $\alpha$ in the range of [0.3, 0.8], and selecting $\alpha=0.5$ guarantees the performance to be within 0.25\% from the peak.

\begin{table*}[!h]
\centering
\resizebox{\textwidth}{!}{
\setlength{\tabcolsep}{0.3em}
\begin{tabular}{ccccccccccccccccc}
 & \rotatebox{90}{Average}        & \rotatebox{90}{ImageNet}       & \rotatebox{90}{ImageNet-A}     & \rotatebox{90}{ImageNet-V2}    & \rotatebox{90}{ImageNet-R}     & \rotatebox{90}{ImageNet-S}         & \rotatebox{90}{Cars} & \rotatebox{90}{Caltech101} & \rotatebox{90}{DTD}  & \rotatebox{90}{EuroSAT} & \rotatebox{90}{FGVC} & \rotatebox{90}{Food101} & \rotatebox{90}{Flower102} & \rotatebox{90}{Pets} & \rotatebox{90}{UCF101} & \rotatebox{90}{SUN397}         \\ \midrule
accuracy std. (\%) & 0.12  & 0.14   & 0.93     & 0.17     & 0.16     & 0.27     & 0.08 & 0.07     & 0.06 & 0.27  & 0.10 & 0.03  & 0.07    & 0.05 & 0.15 & 0.06 \\ \bottomrule \\
\end{tabular}}
\caption{Accuracy standard deviation of Rényi Entropy aggregated predictions over varying $\alpha$.}
\label{std}
\end{table*}

In Table~\ref{std}, we display the accuracy standard deviation of Rényi Entropy aggregated predictions over varying $\alpha$. The table reveals that the accuracy standard deviation is less than 0.3\% on most datasets, with the exception of ImageNet-A (corresponding to the orange curve in Figure~\ref{fig:alpha}). ImageNet-A is composed of challenging outlier examples where machine learning models often falter. It is possible that CLIP produces less confident and flatter prediction logits on ImageNet-A, rendering its performance more sensitive to variations in $\alpha$ compared to other datasets.

\section{Ablation Studies on Prediction Balance Weight $\beta$}
\label{abla:beta}
\begin{table}[ht]
\centering
\begin{tabular}{ccccccccc} \\ 
$\beta$  & 10    & 5    & 4    & 3    & 2    & 1    & 0.5    & 0.1    \\ \midrule
BaFTA & 71.55 & 71.90 & 72.01 & 72.06 & \textbf{72.15} & 71.85 & 71.28 & 70.55 \\ \bottomrule \\
\end{tabular}
\caption{Ablation study on BaFTA performance with different prediction balance weight $\beta$. All results are top-1 accuracy reported with CLIP (ViT-B/16) on ImageNet with 64 augmented views for each test example.}
\label{tab:beta}
\end{table}

In Table~\ref{tab:beta}, we conduct an ablation study to evaluate BaFTA's performance across different values of the prediction balance weight, $\beta$. This parameter $\beta$ determines the balance between text embedding predictions and online clustering predictions in BaFTA's aggregated output, as described in Equation~\ref{eq:renyi2}. A higher $\beta$ value weights the aggregation towards text embedding predictions, while a lower $\beta$ favors online clustering predictions. Our findings reveal that BaFTA reaches optimal performance at $\beta=2$, indicating the importance of having contribution from both text embedding and online clustering predictions. Furthermore, within the range of $\beta\in [1,5]$, BaFTA demonstrates relatively stable performance, deviating by only 0.3\% from its peak. This suggests that BaFTA is robust across a range of $\beta$ values, maintaining efficacy near its best setting.

\section{Ablation Studies on Augmentation Views}
\label{abla:augviews}
\begin{table}[ht]
\centering
\begin{tabular}{cccccc} \\ 
Augmentation Views  & 256    & 128    & 64    & 32    & 16  \\ \midrule
BaFTA & 72.19 & 72.18 & 72.14 & 71.93 & 71.68 \\ \bottomrule \\
\end{tabular}
\caption{Ablation study on BaFTA performance with different number of augmentation views. All results are top-1 accuracy reported with CLIP (ViT-B/16) on ImageNet with 64 augmented views for each test example.}
\label{tab:augviews}
\end{table}

In Table~\ref{tab:augviews} we present the ablation study on BaFTA performance at different number of augmentation sizes. It is shown that BaFTA achieves better performance with an increasing number of augmentation views, but the improvement scale is limited after 64-128 augmented views. We follow the practice from the TPT to use 64 augmentation views to strike the balance of performance and computation costs.  

\section{Effectiveness of Projected Embedding Space}
\label{abla:projection}
\begin{table}[ht]
\resizebox{\textwidth}{!}{
\setlength{\tabcolsep}{0.3em}
\begin{tabular}{ccccccccccccccccc}      \\ 
& \rotatebox{90}{Average}        & \rotatebox{90}{ImageNet}       & \rotatebox{90}{ImageNet-A}     & \rotatebox{90}{ImageNet-V2}    & \rotatebox{90}{ImageNet-R}     & \rotatebox{90}{ImageNet-S}         & \rotatebox{90}{Cars} & \rotatebox{90}{Caltech101} & \rotatebox{90}{DTD}  & \rotatebox{90}{EuroSAT} & \rotatebox{90}{FGVC} & \rotatebox{90}{Food101} & \rotatebox{90}{Flower102} & \rotatebox{90}{Pets} & \rotatebox{90}{UCF101} & \rotatebox{90}{SUN397}         \\ \midrule
$k$NN w/o $P^*$           & 61.56\ & 63.03\ & 44.29\ & 50.56\ & 71.19\ & 44.29\ & 62.49\ & 92.29\ & 43.85\ & 58.37\ & 22.35\ & 68.66\ & 84.56\ & 85.01\ & 68.68\ & 63.77\ \\
$k$NN w/ $P^*$     & 64.04\ & 66.62\ & 48.91\ & 55.41\ & 77.91\ & 47.62\ & 67.01\ & 93.55\ & 45.74\ & 53.75\ & 23.52\ & 69.35\ & 86.33\ & 89.64\ & 69.36\ & 65.87\ \\ \bottomrule \\
\end{tabular}}
\caption{Effectiveness of Projection $P^*$ (Eq.~\ref{proj_eq}) in improving embedding distribution. Results produced with CLIP (ViT-B/16) embeddings, demonstrated by the top-1 accuracy improvement of $k$NN classifier with $k=5$.} 
\label{tab:abla2}
\end{table}

Table~\ref{tab:abla2} provides evidence of the effectiveness of the Projection $P^*$ in enhancing the distribution of CLIP embeddings for clustering, as proposed in \citet{xuefeng2023reclip}. The results demonstrate a 2.48\% improvement in averaged $k$-nearest neighbor (kNN) classifier accuracy across the 15 datasets after projecting the CLIP embeddings with $P^*$. This improvement signifies that $P^*$ successfully enhances the neighboring relationships among CLIP embeddings in the projection space, which, in turn, will benefit the online clustering process.

\begin{figure}[!ht]
\centering
\begin{subfigure}{0.4\textwidth}
  \centering
  \includegraphics[width=\linewidth]{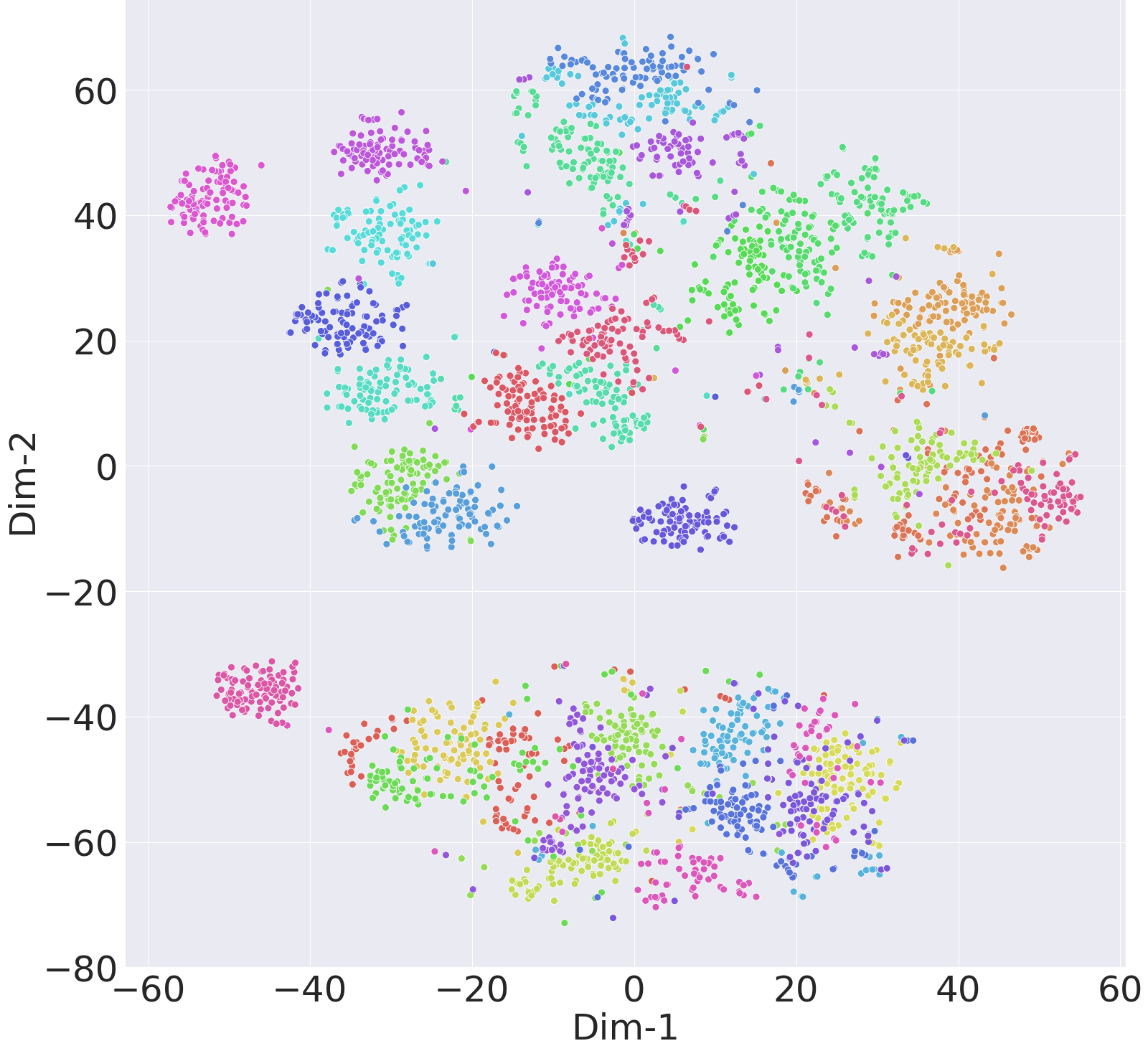}
  \caption{Oxford-IIIT-Pets (Original)}
  \label{fig:sfig1}
\end{subfigure}%
\begin{subfigure}{0.4\textwidth}
  \centering
  \includegraphics[width=\linewidth]{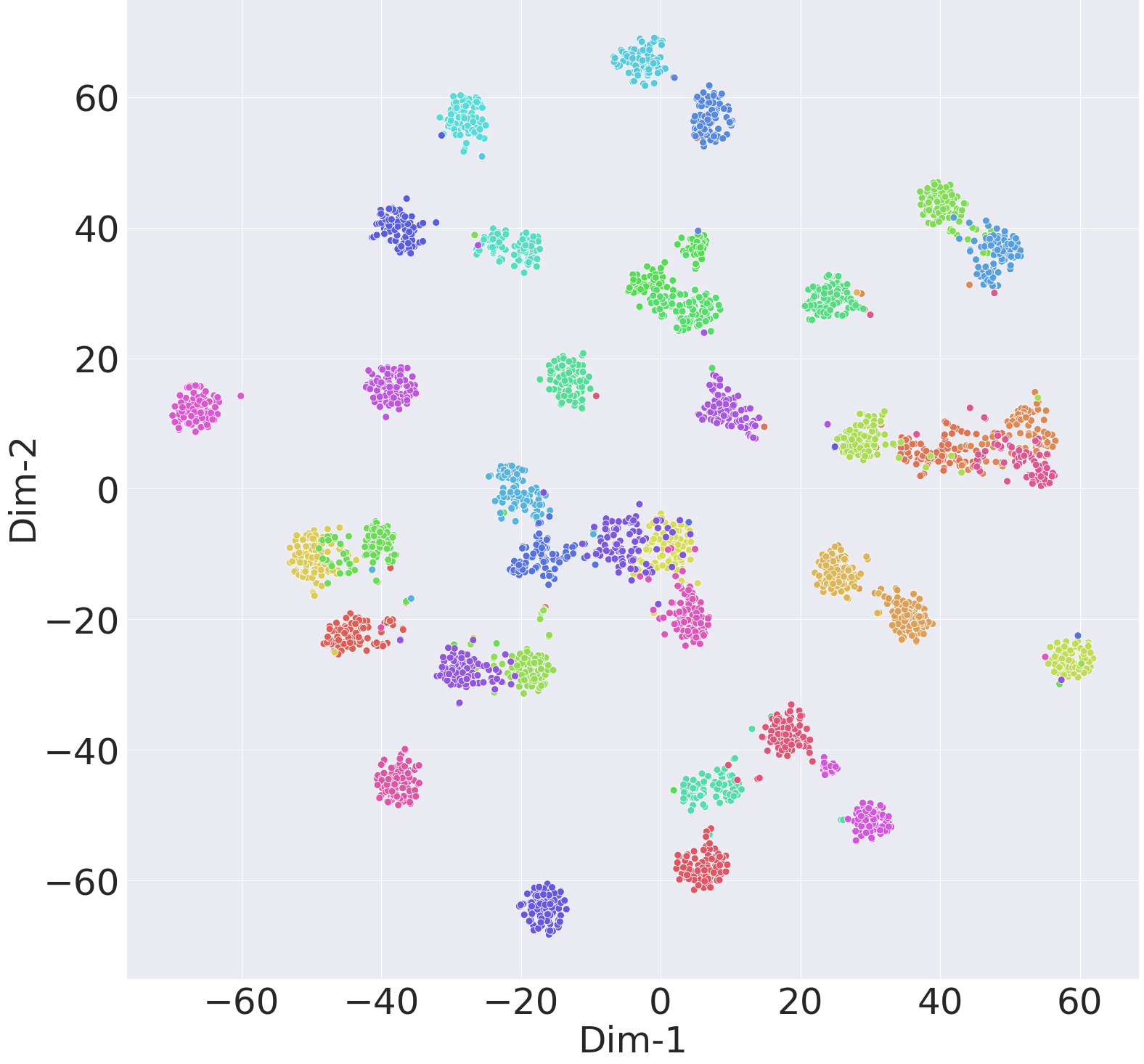}
  \caption{Oxford-IIIT-Pets (Projected)}
  \label{fig:sfig2} 
\end{subfigure} \\ 
\begin{subfigure}{0.4\textwidth}
  \centering
  \includegraphics[width=\linewidth]{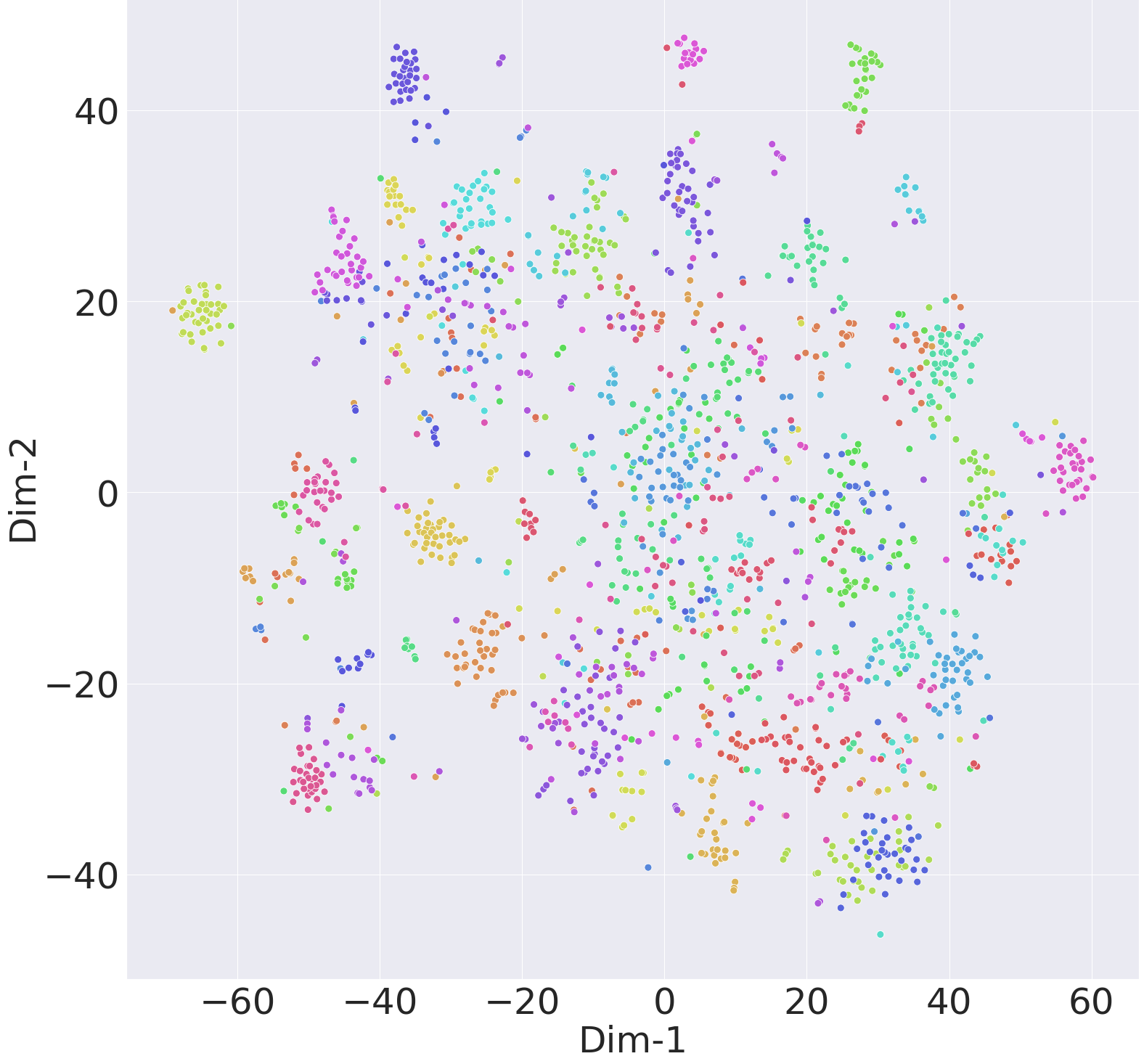}
  \caption{Describable Textures (Original)}
  \label{fig:sfig3}
\end{subfigure}%
\begin{subfigure}{0.4\textwidth}
  \centering
  \includegraphics[width=\linewidth]{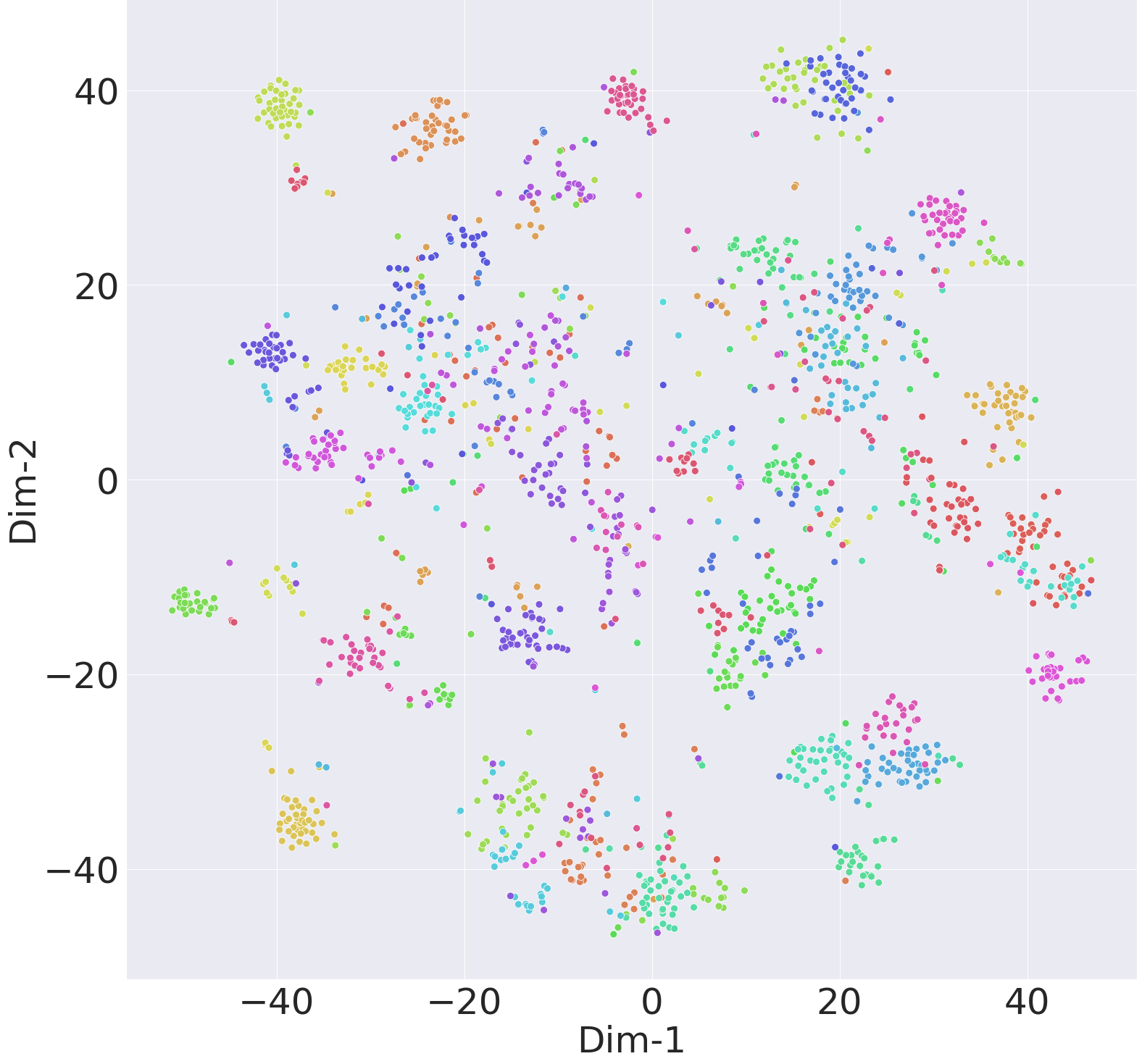}
  \caption{Describable Textures (Projected)}
\label{fig:sfig4} 
\end{subfigure} \\
\begin{subfigure}{0.4\textwidth}
  \centering
  \includegraphics[width=\linewidth]{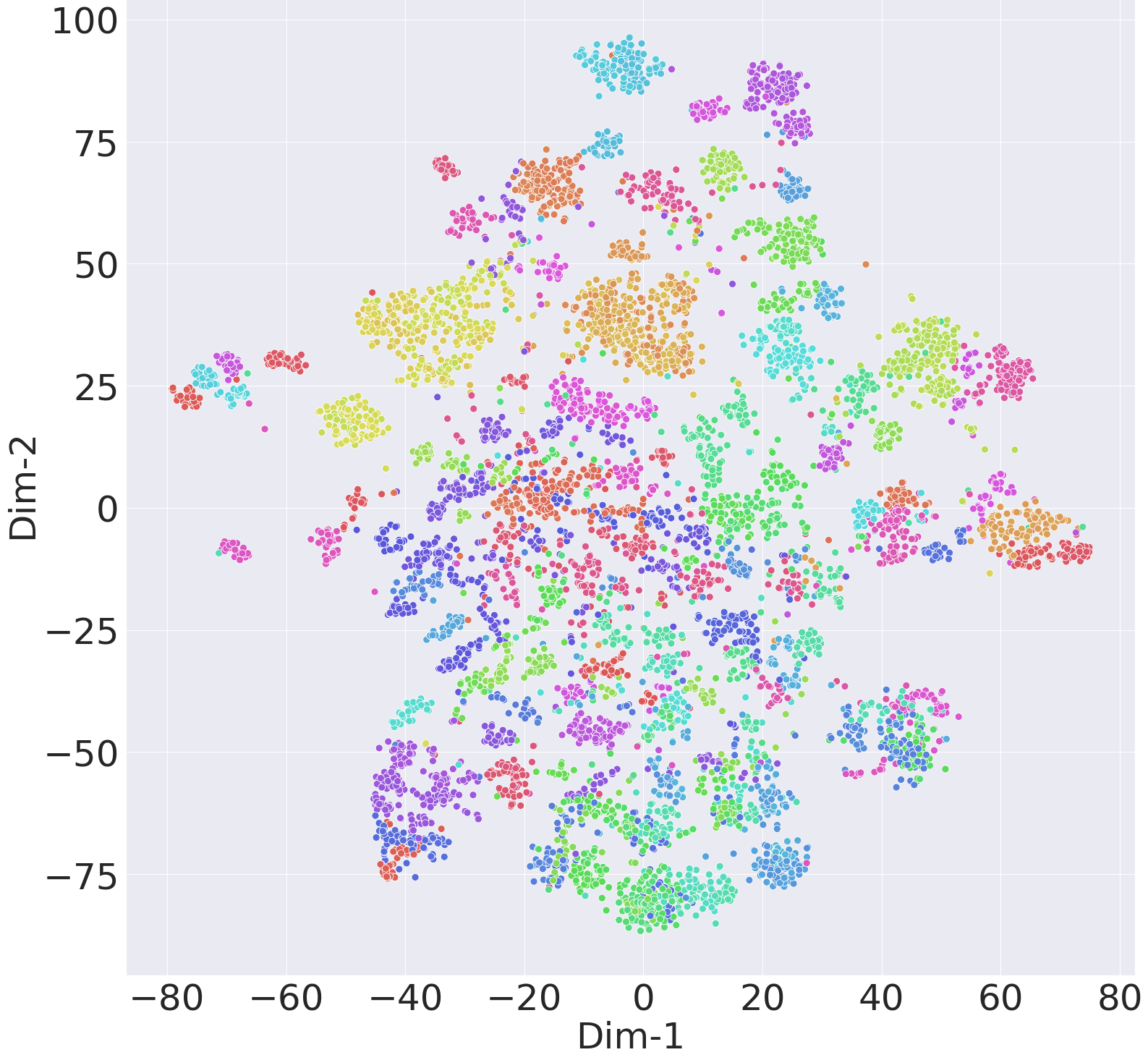}
  \caption{Stanford Cars (Original)}
  \label{fig:sfig5}
\end{subfigure}%
\begin{subfigure}{0.4\textwidth}
  \centering
  \includegraphics[width=\linewidth]{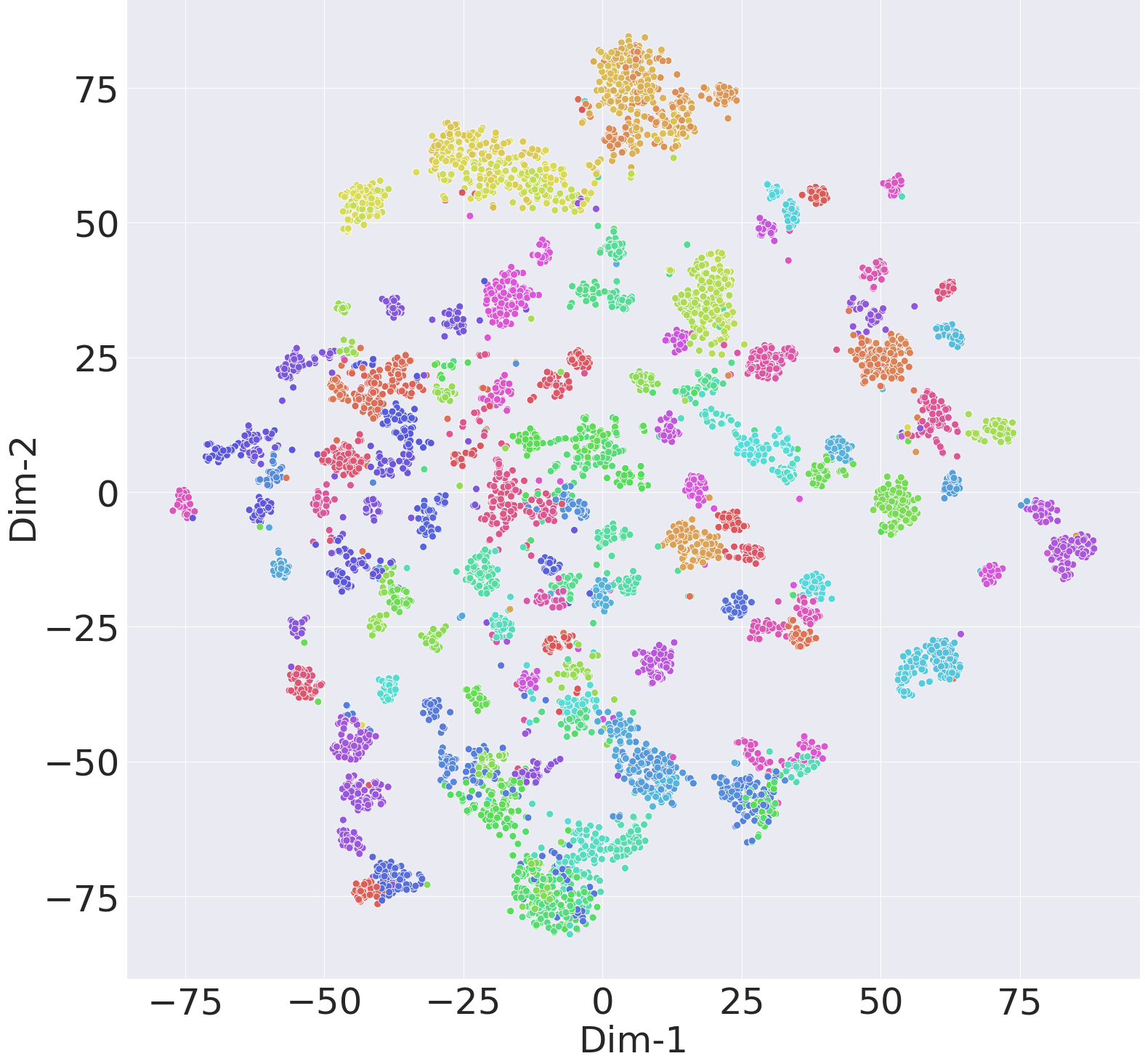}
  \caption{Stanford Cars (Projected)}
\label{fig:sfig6} 
\end{subfigure} \\

\caption{tSNE plots of original and projected visual embeddings from evaluation datasets. }
\label{fig:tsne}
\end{figure}

In Figure~\ref{fig:tsne}, we present t-SNE plots for Oxford-IIIT-Pets, Describable Textures, and Stanford Cars to visually showcase the distribution differences between original CLIP visual embeddings and projected visual embeddings.

As illustrated by the t-SNE plots, the projection space effectively transforms sparse clusters into denser formations, leading to improved online clustering results. Additionally, we observed that the enhancement in clustering brought by the projection is potentially correlated with the classification accuracy on the respective datasets. For instance, CLIP attains a 86.92\% zero-shot accuracy on Oxford-IIIT-Pets, with the projection significantly improving its clustering quality. In contrast, CLIP achieves only 45.04\% accuracy on Describable Textures, and the improvement provided by the projection over the clustering condition is relatively subtle.

\section{Broader Impact}
\label{impact}
The BaFTA algorithm proposes and validates the feasibility of a stable and efficient backpropagation-free method achieving strong overall performance in VLM adaptation, enhancing their zero-shot image classification capability at inference time. This innovation can facilitate the deployment of more efficient and accurate AI systems in various practical applications, such as automated content moderation, medical image analysis, and autonomous driving. By improving the robustness and adaptability of vision-language models without requiring backpropagation or labeled data, BaFTA contributes to reducing computational resources and energy consumption, aligning with environmentally sustainable AI practices. Moreover, the dynamic prediction aggregation mechanism using Rényi entropy improves the prediction reliability across diverse datasets, promoting broader acceptance and trust in AI systems.

\section{Limitations}
\label{limitation}
Despite its notable advantages, BaFTA has certain limitations. Firstly, while it effectively addresses the instability issues of test-time adaptation, its reliance on the quality of initial text embeddings means that it may still struggle with datasets containing ambiguous or less informative class names. Additionally, the algorithm’s performance might degrade in scenarios with highly complex or mixed distributions, where the clustering assumptions do not hold as strongly. Furthermore, BaFTA is primarily designed for zero-shot classification tasks, and its applicability to other tasks, such as image retrieval or multi-modal generation, remains unexplored. Future work should focus on extending BaFTA’s framework to accommodate a broader range of tasks and improving its robustness in handling diverse data distributions.

\end{document}